\PassOptionsToPackage{dvipsnames}{xcolor}
\documentclass{article} %
\usepackage{iclr2027_conference,times}

\usepackage{amsmath,amsfonts,bm}

\def\eqref#1{equation~\ref{#1}}

\def\1{\bm{1}}

\DeclareMathAlphabet{\mathsfit}{\encodingdefault}{\sfdefault}{m}{sl}
\SetMathAlphabet{\mathsfit}{bold}{\encodingdefault}{\sfdefault}{bx}{n}

\usepackage{hyperref}

\usepackage[dvipsnames]{xcolor}
\usepackage{xspace}

\newcommand{\ifprecedingtext}[1]{\ifvmode\relax\else#1\fi}

\newenvironment{redenv}{%
    \color{BrickRed}%
}{%
    \ignorespacesafterend\xspace%
}
\newcommand{\red}[1]{%
    \begin{redenv}#1\end{redenv}  %
}
\newenvironment{blueenv}{%
    \color{blue}%
}{%
    \ignorespacesafterend\xspace%
}

\newenvironment{orangeenv}{%
    \color{orange}%
}{%
    \ignorespacesafterend\xspace%
}

\newenvironment{purpleenv}{%
    \color{Plum}%
}{%
    \ignorespacesafterend\xspace%
}

\newenvironment{oliveenv}{%
    \color{olive}%
}{%
    \ignorespacesafterend\xspace%
}

\newenvironment{greenenv}{%
    \color{ForestGreen}%
}{%
    \ignorespacesafterend\xspace%
}

\newenvironment{bredenv}{%
    \color{red}%
    \bfseries%
}{%
    \ignorespacesafterend\xspace%
    \mdseries%
}

\newcommand{\tocite}[1][]{\red{\if\relax\detokenize{#1}\relax[CITE]\else[CITE #1]\fi}}
\newcommand{\toref}[1][]{\red{\if\relax\detokenize{#1}\relax[REF]\else[REF #1]\fi}}

\newenvironment{tabenv}
   {\list{}{}%
    \item\relax}
   {\endlist}

\usepackage[most]{tcolorbox}    	%
\usepackage{setspace}               
\usepackage{multicol}    
\usepackage{footnotehyper}   %

\definecolor{main}{HTML}{5989cf}    %
\definecolor{sub}{HTML}{cde4ff}     %

\tcbset{
    sharp corners,
    colback = white,
    before skip = 0.2cm,    %
    after skip = 0.5cm      %
}                           %

\newtcolorbox{bbox}{
    colback = sub, 
    colframe = main, 
    boxrule = 0pt, 
    leftrule = 6pt %
}

\definecolor{gboxframe}{HTML}{2E7D32}
\definecolor{gboxback}{HTML}{EDF7ED}
\newtcolorbox{gboxenv}{
  colback=gboxback,
  colframe=gboxframe,
  boxrule=1.2pt,
  arc=8pt,
  outer arc=8pt,
  rounded corners,
  left=6pt, right=6pt, top=3pt, bottom=3pt %
}
\makesavenoteenv{gboxenv}

\usepackage{url}
\usepackage{graphicx}
\usepackage{enumitem}
\usepackage{xcolor}

\title{COGENT: A Modular Cognitive Architecture\\for Agentic User Simulation}

\author{Yikun Chi$^1$\thanks{Work completed during an internship at Microsoft.}
\quad Adam Davies$^2$ \quad Debarati Das$^2$ \quad Lin Ai$^2$ \quad Scott Counts$^2$ \\
$^1$Stanford University \quad $^2$Microsoft
}

\iclrfinalcopy
\hypersetup{
  pdftitle={COGENT: A Modular Cognitive Architecture for Agentic User Simulation},
  pdfauthor={Yikun Chi, Adam Davies, Debarati Das, Lin Ai, Scott Counts}
}
\usepackage[T1]{fontenc}
\usepackage{graphicx}
\usepackage{booktabs}
\usepackage{array}
\usepackage{tabularx}
\usepackage{longtable}
\usepackage{amsmath}
\usepackage{amssymb}
\usepackage{microtype}
\usepackage{xspace}
\usepackage{placeins}
\newcommand{\cogent}{\textsc{Cogent}\xspace}
\newcommand{\cua}{\textsc{CUA}\xspace}
\newcommand{\cuauser}{\textsc{CUA-User}\xspace}
\graphicspath{{draft/figures/}}

\begin{document}
\maketitle
\lhead{Preprint}
\begin{abstract}
LLM-based user simulation can support interface design, accessibility research, and interactive-agent evaluation, but standard computer-use agents are optimized for task completion rather than reproducing how people perceive, reason, and act. We introduce \cogent (\textbf{COG}nitive ag\textbf{ENT}), a modular cognitive architecture for simulating user interaction with desktop interfaces. \cogent draws on established cognitive theories, such as Norman's action cycle and ACT-R, to enable explainable, intervenable, and modular user simulation by decomposing the simulation process into cognitive modules with explicit representations of perception, memory, user state, intentions, and action selection. 
We begin by presenting six intervenability demonstrations, each tracing how a single edit to memory, a user attribute, or a processing module propagates through the cognitive pipeline to produce a behavioral change. 
We evaluate \cogent's behavioral and attributional fidelity across three benchmarks -- AgentNetBench, A11y-CUA, and SENSE-42 -- finding that
it achieves statistically significant fidelity gains over computer-use baselines in reproducing recorded human behavior and recovers the relational structure among human self-reported psychological states. 
Together, these demos and evaluations  
demonstrate \cogent's utility toward more faithful, explainable, intervenable, and modular LLM-based cognitive simulations.

\end{abstract}

\section{Introduction}
\label{sec:introduction}
Large language models (LLMs) are increasingly used to simulate people. Commercial platforms such as Aaru, Artificial Societies, Simile, and Viewpoints.ai offer population, audience, and user simulations \citep{aaru2026simulation,artificialsocieties2026simulation,simile2026simulation,viewpoints2026simulation}. Academic work uses LLMs to represent survey respondents, students, economic actors, and agents participating in extended social interaction \citep{argyle2023out,horton2023large,park2023generative,zhou2024sotopia,mannekote2025can,mannekote2025do}. In $\tau$-bench and UserBench, AI components also role-play users to evaluate another agent \citep{yao2024taubenchbenchmarktoolagentuserinteraction,qian2025userbench}. These applications raise a common question: \textit{what aspects of a person should a simulated user reproduce?}

\paragraph{Reaching the goal is not the same as behaving like a user.}  People can reach the same outcome through different routes. One recalls a shortcut, another searches for a visible control, and another asks for help after an unsuccessful attempt. Their choices depend on what they notice, what they know, how they interpret the situation, and how their state changes with experience. Computer-use agents are optimized for the outcome, so they leave the route unmodeled. Recent comparisons with human users report exactly this failure: simulated users diverge from real ones, and the evaluations built on them inherit the divergence \citep{seshadri2026lost,zhou2026mindsim2realgapuser}. For interface research and cognitive science, the route is the object of study, not a nuisance.

\paragraph{Prompted personas expose no mechanism to study.} Asking a model to ``act frustrated'' produces text that reports frustration, but nothing connects that report to behavior. \citet{goldenberg2026emotions} argue for a functional account of emotion, one defined by its role in subsequent behavior. A simulation of frustration therefore needs an explicit pathway: experience updates the state, and the state changes persistence or action. Memory needs the same treatment, affecting recognition and choice rather than sitting in a prompt. A monolithic prompt offers researchers no place to intervene, no intermediate representation to inspect, and no way to attribute a behavioral change to a specific assumption. We argue that user simulation needs an architecture, not a longer persona.

\paragraph{\cogent.} We introduce \cogent, a cognitive architecture for agentic user simulation. Its design follows three principles: \emph{explainability} through traceable intermediate representations, \emph{intervenability} through edits to memory, attributes, and processing modules, and \emph{modularity} through reusable components with declared interfaces. Inspired by ACT-R and Norman's action cycle, five modules connect screenshots to cognitive processing, control, action selection, and execution. Shared memory, a situation model, and user attributes maintain continuity across interactions.

\paragraph{Contributions.}
\begin{enumerate}[leftmargin=*,itemsep=2pt,topsep=2pt]
\item \textbf{An architecture that makes cognitive assumptions editable.} \cogent{} exposes memory, user attributes, and individual processing modules as separate intervention points, with typed records connecting an edit to the behavior it produces.
\item \textbf{Six paired demonstrations of intervention.} Two edits to memory, two to user attributes, and two to processing modules. Each pair traces a single change through the pipeline to a recorded difference in behavior, showing what an architecture buys that a prompt does not.
\item \textbf{Behavioral fidelity against human reference data.} On AgentNetBench \citep{wang2025opencuaopenfoundationscomputeruse}, \cogent{} predicts 59.8\% of recorded human actions exactly, against 53.7\% for a computer-use agent prompted to simulate a user (paired gain 6.04 points, 95\% CI [2.99, 9.20]). On A11y-CUA \citep{Gubbi_Mohanbabu2026-wn}, it yields lower action-sequence distance across 43 matched tasks (0.511 versus 0.612). Fidelity is measured three ways: agreement with the recorded human choice, similarity of action-type distributions, and similarity of action sequences.
\item \textbf{Attribute fidelity and its limits.} Using SENSE-42 \citep{Zhang2026-qn}, we compare relationships among internal states, and their associations with behavior, against human self-reports. The agent recovers much of the relational structure ($\rho_s = 0.860$) while their magnitudes and their links to behavior remain imperfectly calibrated. We report this negative result rather than omit it.
\end{enumerate}

\section{Related Work}
\label{sec:related}

\paragraph{LLMs as human simulations.}
Research on LLM-based human simulation spans a growing range of behavioral settings, from eliciting survey responses to sustaining interaction in social and task environments. One line of work examines whether conditioning a model on demographic characteristics, preferences, or situational information can reproduce human judgments and choices. Studies of synthetic survey respondents, economic agents, and consumer samples approach simulation at this level \citep{argyle2023out,horton2023large,sarstedt2024using}. Interactive simulations introduce an additional requirement: behavior must develop through experience and exchanges with others. Generative Agents uses memory, reflection, and planning to support continuing social activity, while Sotopia provides environments for evaluating goal-directed social interaction \citep{park2023generative,zhou2024sotopia}. Simulated users also serve as interaction partners in agent benchmarks, where their responses shape another agent's opportunities to act \citep{yao2024taubenchbenchmarktoolagentuserinteraction,barres2025tau,qian2025userbench}. Across these settings, a central question is which aspects of human behavior the simulation preserves. Reported discrepancies between simulated and human users, including differences in the evaluations they produce, motivate validation against human observations \citep{seshadri2026lost,zhou2026mindsim2realgapuser}. For sustained interaction, this also calls for examining how knowledge and internal states influence behavior as a situation unfolds.

\paragraph{LLMs and computer use.}
Computer-use agents are progressing toward autonomous interaction across web and desktop environments through advances in visual grounding, intermediate reasoning, and learning from demonstrations \citep{qin2025uitarspioneeringautomatedgui,wang2025opencuaopenfoundationscomputeruse}. Existing agents already incorporate processes associated with deliberate human problem solving: UI-TARS uses task decomposition, reflection, and milestone recognition within a single model \citep{qin2025uitarspioneeringautomatedgui}. Alongside this integrated approach, modular architectures distribute responsibilities among components. Agent S2 develops this direction by combining generalist planning with specialist grounding \citep{agashe2025agents2}.

In parallel, benchmarks have expanded from predicting actions on websites, as in Mind2Web, to executing multi-step tasks in realistic web environments and workflows spanning desktop applications, as in WebArena and OSWorld \citep{deng2023mind2webgeneralistagentweb,zhou2024webarenarealisticwebenvironment,xie2024osworldbenchmarkingmultimodalagents}. Successful task completion remains the central objective of these interactive benchmarks. Human-referenced evaluation also examines efficiency in OSWorld-Human and accessibility-related interaction patterns in A11y-CUA \citep{abhyankar2025osworldhuman,Gubbi_Mohanbabu2026-wn}. Fidelity to human behavior receives less attention than task competence. For user simulation, the growing capabilities of computer-use agents therefore need to be accompanied by evaluation of how their action choices and trajectories correspond to those of people, including variation associated with knowledge, strategy, and user state.

\paragraph{Cognitive architectures and interaction.}
A substantial body of cognitive-architecture research supports modeling cognition through distinct, interacting components. ACT-R makes memory, goals, and perceptual-motor processing explicit, while the Standard Model of the Mind synthesizes shared commitments across ACT-R, Soar, and Sigma, including distinctions among working memory, long-term memory, perception, and motor systems \citep{anderson2004integrated,laird2017standard}. Such decompositions connect behavioral predictions to specific assumptions about how information is processed. Within human--computer interaction, Norman's action cycle relates goals and action execution to the interpretation of feedback \citep{norman1986cognitive}. Accounts of display-based action, exploratory learning, and cognitive walkthroughs further explain how visible cues and prior knowledge guide interface use, while soft-constraints theory examines the tradeoff between relying on memory and consulting the environment \citep{payne1991display,polson1990theory,rieman1996field,wharton1994cognitive,gray2004soft}. This tradition is now informing language-agent design: CoALA organizes memory, internal and external actions, and decision making within a common framework, and brain-inspired architectures apply process decomposition to planning \citep{sumers2024cognitive,Webb2025-xe}. Together, these approaches provide a foundation for constructing agents whose behavior arises through inspectable interactions among cognitive processes. COGENT applies this perspective to user simulation, making perception, memory, user state, and action selection explicit and editable within desktop interaction.

\section{COGENT's Design}
\label{sec:design}

{
Prior LLM-agent architectures establish important foundations for modular reasoning, memory, reflection, and user simulation, from general frameworks such as CoALA \citep{SumersCog} and Concordia \citep{vezhnevets2023generative} to usability-oriented systems such as SimUser \citep{xiang2024simuser} and UXAgent\citep{lu2025uxagent}. Building on these foundations,  COGENT focuses on modeling how cognitive processes shape human interaction behavior, with an emphasis on making these processes explicit, inspectable, and configurable.
Specifically, we followed three design principles. \emph{Explainability} means that downstream modules use upstream artifacts in a traceable way, connecting model assumptions to behavior \citep{Epstein1999,anderson2004integrated}. \emph{Intervenability} supports changing those assumptions and examining their consequences \citep{epstein2008why}: researchers can edit memory, configure attributes, or replace processing modules. \emph{Modularity} supports assembling users from reusable components with declared interfaces, following component-based generative simulation systems such as Concordia \citep{vezhnevets2023generative}.
}

\begin{figure}[tbp]
\centering
\includegraphics[width=0.95\textwidth]{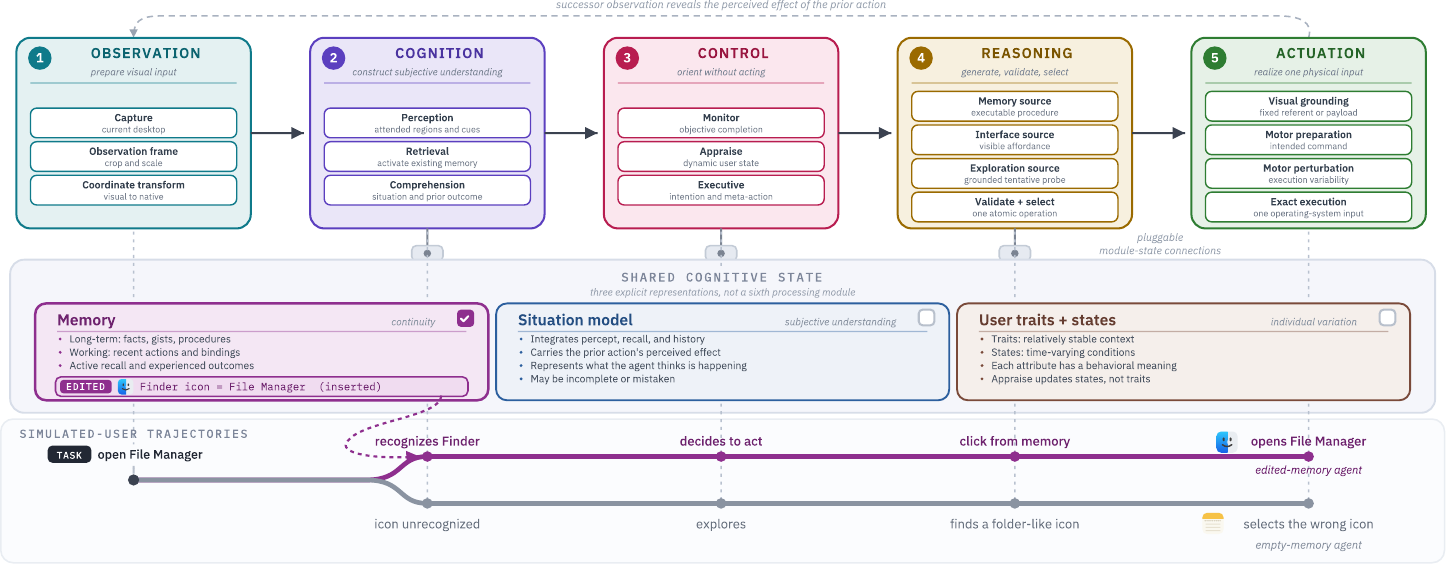}
\caption{\textbf{COGENT's architecture.} Five modules coordinate through memory, a situation model, and user attributes. Actuation denotes Execution. The lower schematic traces a memory intervention through the processing cycle. 
}
\label{fig:architecture}
\end{figure}

\paragraph{Cognitive inspiration and processing cycle.}
Two theories inspired the architecture. ACT-R motivates specialized processes exchanging information through explicit representations; Norman's action cycle organizes the progression from intention to action and interpretation of its consequences \citep{anderson2004integrated,norman1986cognitive}. \emph{Observation} captures the interface. Within \emph{Cognition}, Perception identifies noticed regions and produces a subjective image, Retrieval activates stored knowledge, and Comprehension constructs a situation model. Within \emph{Control}, Monitor assesses completion, Appraise updates dynamic states, and Executive maintains an intention or chooses to rest, request help, or give up. Appraisal accounts motivate relating state updates to interpreted experience \citep{scherer2005emotions}. \emph{Action Reasoning} generates candidates from remembered procedures, perceived controls, and exploratory probes, retaining their sources and expected outcomes. \emph{Execution} grounds the selected referent and dispatches the fixed mouse or keyboard operation; the next observation supplies feedback.

Three shared stores connect cycles: \emph{memory} holds facts, procedures, experiential summaries, and bounded working contents; the \emph{situation model} holds the user's current interpretation; and \emph{attributes} hold stable traits and dynamic states, each with a meaning and value. Their roles draw on accounts of memory access and capacity, situation comprehension, and goal maintenance \citep{cowan2001magical,zwaan1998situation,miller2001integrative}. The five-module grouping and LLM implementations realize these functions through typed records. Appendix~\ref{app:design-details}, particularly Appendix~\ref{app:design-processes}, details the theoretical motivations and processing operations.

\paragraph{Information boundaries.}
Each process receives preceding-stage outputs and the shared-store information declared for its role. Retrieval identifies existing memory items. Comprehension builds its situation from subjective percepts, recalls, and recent experience; Control evaluates that interpretation and updates intentions and states. Action candidates refer to noticed controls or retain recalled procedure templates. Among downstream modules, Execution alone receives the original screenshot for locating the already selected referent; checks preserve the operation and payload. These boundaries make an intervention's downstream path traceable while separating selection from physical grounding. Pretrained knowledge also contributes to interpretation, motivating the memory studies discussed in Section~\ref{sec:discussion}. Appendix~\ref{app:design-boundaries} specifies the module interfaces.

\section{Intervenability Demonstrations}
\label{sec:atomic}

\begin{figure}[!t]
\centering
\includegraphics[width=\linewidth]{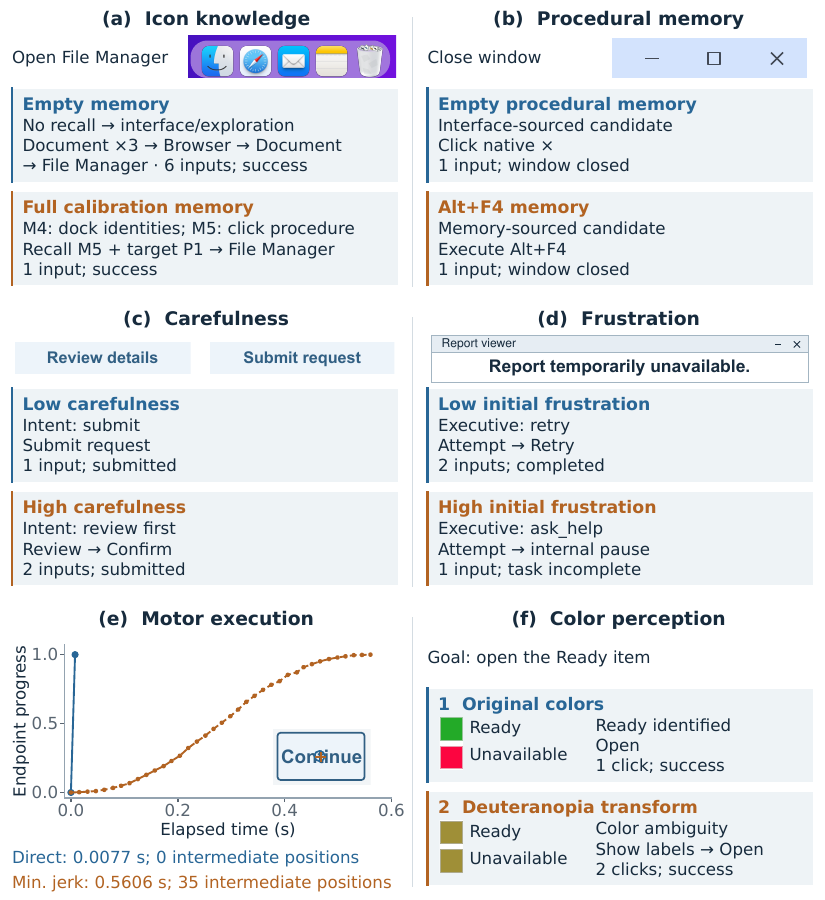}
\caption{\textbf{Six paired interventions.} (a) Empty versus full calibration memory changes icon search. (b) A recalled procedure changes window closing from a click to Alt+F4. (c) Carefulness changes review behavior. (d) Frustration changes persistence and the Executive's help-seeking decision. (e) Replacing the motor process changes recorded movement (where the blue circle marks the direct-positioning endpoint, and the orange plus marks the minimum-jerk click endpoint). (f) Original color input enables direct selection; a deuteranopia transform prompts label revelation first. Crops and cursor samples come from recorded evidence; text summarizes intermediate records and executed behavior.}
\label{fig:atomic}
\end{figure}

Researchers can configure a user simulation by changing the environment in which the agent acts or by changing the simulated user. For the latter, \cogent provides three distinct forms of intervention: editing memory, setting user attributes, and replacing processing modules. We present six paired intervenability demonstrations, two for each form, to show how these edits propagate through the architecture and produce differences in behavior (Figure~\ref{fig:atomic}). Each pair contains one recorded trajectory per condition, with intermediate records connecting the intervention to the agent's actions and the task outcome.

\paragraph{Memory: Configure what the user knows.}
Memory configuration specifies the knowledge available to the simulated user, including semantic knowledge about interface elements and procedural knowledge about how to act. The two demonstrations illustrate how these forms of knowledge affect recognition and the choice of an action.
\emph{Demo (a): icon knowledge} shows how prior knowledge changes search in an unfamiliar dock. With empty memory, the agent first clicks Document, then tries other dock icons before opening File Manager after six inputs. With full calibration memory, containing semantic knowledge about the icons and an executable procedure, the agent recognizes File Manager and opens it on the first click.
\emph{Demo (b): procedural memory} shows how a remembered procedure changes the route to the same outcome. With empty memory, the agent clicks the window's visible close button. With a stored Alt+F4 procedure, it recalls and executes the keyboard shortcut. Both conditions close the window with one input.

\paragraph{Attributes: Configure traits and states.}
User attributes represent traits held fixed during a simulation and states updated as interaction unfolds. Researchers can edit a trait to change a stable behavioral tendency or set an initial state to change how the user enters a situation. The two demonstrations edit carefulness and initial frustration, respectively.
\emph{Demo (c): carefulness} shows how a stable trait changes whether the agent checks its work before submission. With low carefulness, the agent submits the prepared appointment request directly. With high carefulness, it reviews the request before confirming submission. Both conditions succeed, using one and two inputs, respectively.
\emph{Demo (d): frustration} shows how an initial state changes persistence after an unsuccessful attempt. With low initial frustration, the agent retries access to an initially unavailable report and completes the task after two inputs. With high initial frustration, it attempts access once and then the Executive selects \texttt{ask\_help}, pausing the task. This help-seeking decision is recorded internally and produces no additional physical input.

\paragraph{Modules: Replace individual processes.}
Researchers can replace processing modules to implement alternative accounts of perception, appraisal, action selection, or execution. This allows a simulation to incorporate a specific model of how a user processes information or performs an action. The following demonstrations illustrate this flexibility through changes to motor execution and perceptual input.
\emph{Demo (e): motor execution} shows how replacing the motor process changes the movement used to carry out the same intention. With direct positioning, the cursor reaches the Continue button in 0.0077 seconds, with no sampled intermediate positions. With a minimum-jerk process, it moves through 35 distinct intermediate positions over 0.5606 seconds. Both conditions click the intended button.
\emph{Demo (f): color perception} shows how changing perceptual input changes the strategy used to identify a target. With the original screenshot, the agent distinguishes green Ready items from red Unavailable items and opens the Ready item in one click. With a deuteranopia-transformed screenshot, the status colors are difficult to distinguish, so the agent first clicks \emph{Show status labels}, then uses the revealed text to identify and open Ready with a second click. Both conditions succeed through different information-seeking behavior.

The details of the demo can be found in Appendix~\ref{app:demos}.  These demonstrations illustrate how researchers can edit a simulated user in individual scenarios. More generally, in Appendix~\ref{sec:ablation}, we examine how such module edits affect simulated behavior at the distribution-level across a broader set of tasks.

\section{Experiments}
\label{sec:experiments}

We evaluate \emph{behavioral fidelity} against human choices and interaction sequences, and \emph{attribute fidelity} against human-reported states and their behavioral relationships. Behavioral evaluations compare \cogent with a direct computer-use agent (\cua),  a baseline prompted to simulate a human user (\cuauser), {and a baseline with a fixed model-call budget (CUA-User-CB). CUA-User-CB generates five independent action proposals using the same prompt as \cuauser and the recent executed actions. This budget approximately matches \cogent based on overall step and model-call counts on A11y-CUA. A majority vote decides on the final action, with ties resolved by first appearance and no additional model call.} All use GPT-5.4; their prompts, intermediate representations, and model-call counts differ (Appendix~\ref{app:agent-comparison}).

\paragraph{AgentNetBench.}
On 100 OpenCUA/AgentNet tasks, agents predict 1,208 recorded human actions from screenshots and preceding context \citep{wang2025opencuaopenfoundationscomputeruse}. Human actions determine subsequent states, holding the observation sequence fixed. Exact-action agreement checks the normalized operation and relevant keys or text; pointer actions must fall within an accepted target region. Action-type agreement uses 1,111 targets mapping to the eight physical-action classes. Direct baselines retain screenshot/action context, whereas COGENT retains cognitive state. COGENT and \cuauser receive the same fixed participant-context description. An annotated subset additionally distinguishes choosing a workable alternative from matching the recorded human choice. Appendix~\ref{app:agentnet} gives scoring, alternative-action annotations, and full results (Table~\ref{tab:agentnet-full}).

\paragraph{A11y-CUA.}
We evaluate 43 Windows tasks with a 30-step budget and independent outcome review \citep{Gubbi_Mohanbabu2026-wn}. The main comparison uses the tasks for all four agents, including successful, failed, and unverifiable attempts: 172 agent trajectories and 339 successful sighted-human references. Appendix~\ref{app:a11y} reports cohort analyses for the original three agents; task selection and outcome assessment appear in Appendix~\ref{app:a11y-archive}, and the full ledger in Appendix~\ref{app:a11y-ledger} (Table~\ref{tab:a11y-task-ledger}).

\paragraph{Metrics and uncertainty.}
Both behavioral benchmarks use eight physical-action classes. Jensen--Shannon (JS) distance compares pooled action counts. A11y sequence distance uses normalized Levenshtein distance over the common prefix of each agent--human pair; comparisons average humans within task and then tasks equally. Let $D^{AH}_{a,t}$ denote this agent--human distance and $D^{HH}_t$ the mean distance between distinct human references. The human-normalized gap is\footnote{
    The value of $10^{-6}$ in the denominator is arbitrary; its only purpose is to avoid dividing by zero.
}
\begin{equation}
\operatorname{HNG}_{a,t}=\frac{D^{AH}_{a,t}}{D^{HH}_t+10^{-6}}.
\label{eq:hng}
\end{equation}
A value of one denotes average human--human disagreement. HNG weights deviations relative to each task's human variability. Paired task-bootstrap comparisons use nominal 95\% intervals, unadjusted across benchmark contrasts. Appendix~\ref{app:fidelity-metrics} specifies normalization, resampling, and distance definitions (Equations~\ref{eq:js} and~\ref{eq:sequence}); trajectory lengths accompany matched-prefix comparisons.

\paragraph{SENSE-42.}
We compare 42 participants and 1,078 task--questionnaire pairs with one partial simulation run containing 55 substantive tasks and 750 state updates \citep{Zhang2026-qn}. Seven attributes cover sleepiness, mental and temporal demand, perceived performance, effort, frustration, and attentiveness. Attribute meanings are held fixed; Appraise generates updates from task experience. Memory and dynamic states persist across task boundaries, while objectives and task-local buffers reset. We compare the 21 pairwise attribute correlations using rank agreement, magnitude error, and an exact attribute-label permutation test. Each human relationship is the median within-participant correlation, compared with the corresponding simulator correlation. Exploratory hypotheses relate temporal and mental demand to pointer--keyboard switching and effort to decisions per scored unit. Human support requires a predicted-direction association with a familywise 98.33\% interval excluding zero before examining the agent association. Human intervals resample participants; agent intervals resample tasks within the single run. Appendix~\ref{app:sense} provides the protocol, including task sequence, memory, resets, and scales in Appendix~\ref{app:sense-protocol}.

Appendix~\ref{app:provenance} describes the data and materials for reproducing these analyses.

\section{Results}
\label{sec:results}

\subsection{Behavioral fidelity}

Across the two behavioral benchmarks, \cogent more closely matches human action choices and interaction patterns than \cua or \cuauser.

On AgentNetBench, \cogent achieves significantly higher next-action prediction accuracy and lower action-distribution distance than both baselines (Table~\ref{tab:agentnet}). Exact-action agreement is 59.8\%, compared with 52.2\% for \cua and 53.7\% for \cuauser. Action-type accuracy is also higher, at 84.1\% versus 79.0\% and 80.9\%, respectively. All four prediction gains are statistically significant.

The predicted action-type distribution is also closer to the human distribution: pooled JS distance is 0.0445 for \cogent, compared with 0.1252 for \cua and 0.1210 for \cuauser. Both reductions are statistically significant. Together, these results suggest that \cogent better predicts both a user's specific next action and the overall mix of actions users choose across recorded interface states. Appendix~\ref{app:agentnet} contains more detailed analysis.  

\begin{table}[htbp]
\centering
\caption{\textbf{AgentNetBench: human-action prediction and action distributions.} Accuracies are percentages; JS compares pooled action counts. Subscripts give 95\% task-bootstrap CIs. Bold marks significant improvements over both baselines in paired comparisons. Exact/type denominators are 1,208/1,111.}
\label{tab:agentnet}
\small\begin{tabular}{@{}lcc@{}}
\toprule
Agent & Exact action $\uparrow$ & Action type $\uparrow$ \\
\midrule
COGENT & \textbf{59.8}$_{[56.4, 63.0]}$ & \textbf{84.1}$_{[81.6, 86.5]}$ \\
CUA & 52.2$_{[48.8, 55.3]}$ & 79.0$_{[76.1, 81.9]}$ \\
CUA-User & 53.7$_{[50.3, 57.1]}$ & 80.9$_{[78.2, 83.6]}$ \\
\bottomrule
\end{tabular}

\end{table}

A11y-CUA reproduces the action-distribution advantage observed on AgentNetBench and extends it to action sequences during live interaction (Table~\ref{tab:a11y-main}). Across 43 tasks, pooled JS distance is 0.272 for \cogent, compared with 0.370 for \cua,  0.418 for \cuauser and 0.389 for CUA-User-CB; all reductions are statistically significant. Sequence distance is also significantly lower against \cua and \cuauser, at 0.511 versus 0.595 and 0.612, respectively. However, CUA-User-CB, a persona baseline with an approximately matched compute budget, has no statistically significant sequence-distance difference compared to \cogent. These findings suggest that \cogent more closely reproduces the relative frequencies of human action types.

\begin{table}[htbp]
\centering
\caption{\textbf{A11y-CUA: 43 matched tasks, including all outcomes.} Lower is closer to humans. Subscripts give 95\% task-bootstrap CIs; bold marks significant improvements over \cua and \cuauser in paired comparisons. JS pools actions; sequence distance and HNG weight tasks equally.}
\label{tab:a11y-main}
\small{
\begin{tabular}{@{}lrccc@{}}
\toprule
Agent & Tasks & JS distance $\downarrow$ & Sequence distance $\downarrow$ & HNG $\downarrow$ \\
\midrule
COGENT & 43 & \textbf{0.272}$_{[0.239, 0.320]}$ & \textbf{0.511}$_{[0.467, 0.557]}$ & 2.020$_{[1.572, 2.546]}$ \\
CUA & 43 & 0.370$_{[0.330, 0.432]}$ & 0.595$_{[0.527, 0.660]}$ & 2.198$_{[1.761, 2.682]}$ \\
CUA-User & 43 & 0.418$_{[0.362, 0.493]}$ & 0.612$_{[0.546, 0.681]}$ & 2.153$_{[1.814, 2.506]}$ \\
CUA-User-CB & 43 & 0.389$_{[0.351, 0.437]}$ & 0.532$_{[0.478, 0.580]}$ & 1.890$_{[1.577, 2.280]}$ \\
\bottomrule
\end{tabular}
}

\end{table}

After normalizing by within-task human variability, however, the differences in HNG are not statistically significant. Mean HNG values above one indicate that agent--human disagreement remains substantially larger than human--human disagreement. HNG gives greater weight to deviations on tasks where humans behave consistently, so gains in raw sequence distance need not translate into significant normalized gains.

Appendix ~\ref{app:a11y} contains additional details of the analysis.

\subsection{Attribute fidelity}

In SENSE-42, \cogent captures much of the ordering of relationships among user attributes, while their magnitudes remain imperfectly calibrated. The tested attribute--behavior associations in the agent are not statistically significant.

\begin{figure}[htbp]
\centering
\includegraphics[width=.5\linewidth]{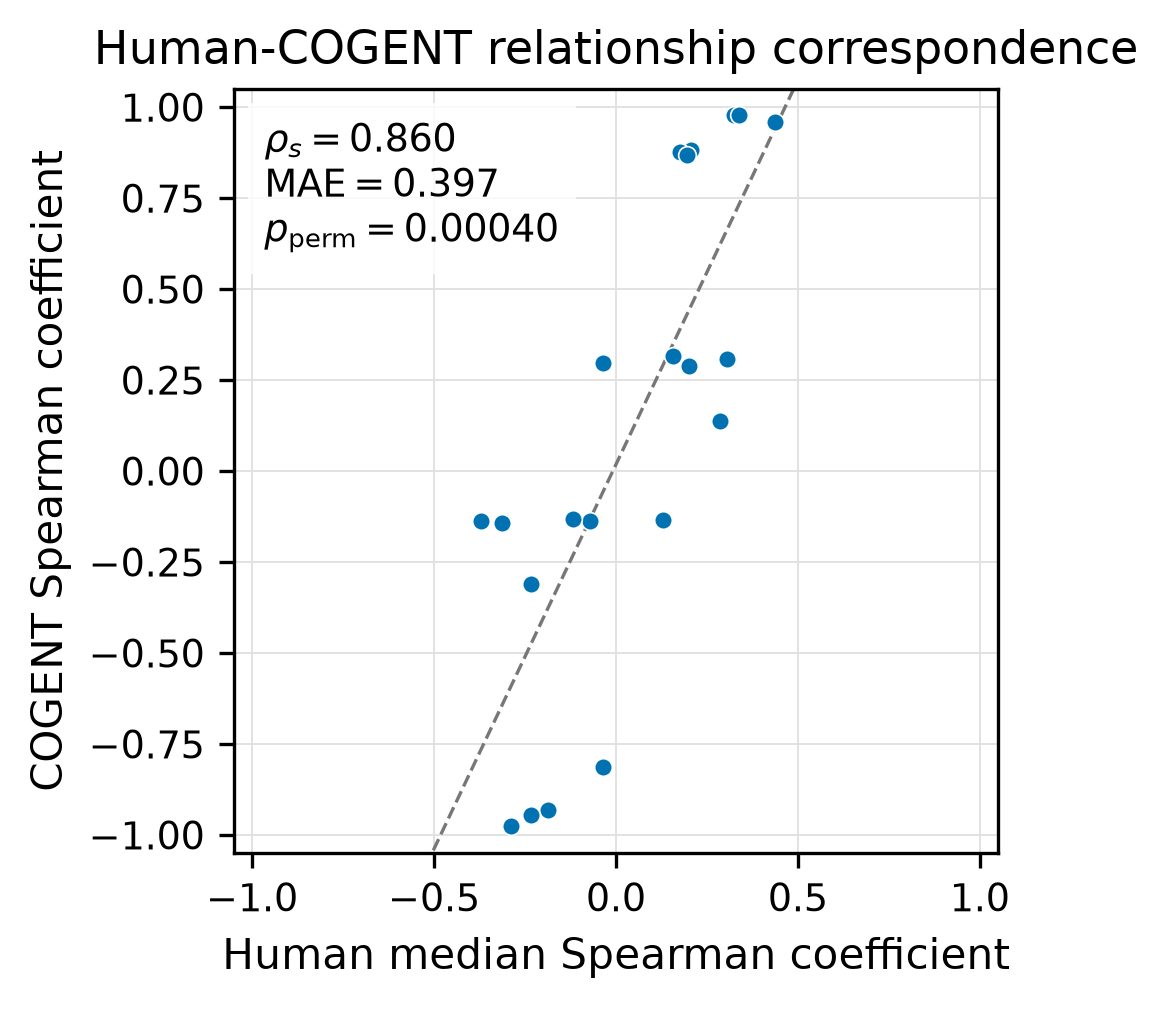}
\caption{SENSE-42 relationship-structure comparison. Each point represents one of 21 attribute pairs: the human coordinate is a median of within-participant correlations, and the agent coordinate uses 750 updates from one simulation run. Rank agreement is $\rho_s=0.860$ (exact attribute-label permutation $p=0.000397$). The dashed line is a descriptive fit.}
\label{fig:sense}
\end{figure}

\paragraph{Relational structure and calibration.}
The ordering of 21 pairwise attribute relationships shows strong agreement with the human pattern ($\rho_s=0.860$), and this agreement is statistically significant (Figure~\ref{fig:sense}). Correlation-magnitude MAE is 0.397. Thus, the agent largely preserves which attribute relationships are relatively stronger or weaker, but the strength of those relationships remains a calibration target. Appendix~\ref{app:sense-structure} gives the 55-task-mean sensitivity, distinguishing task means from the 750 updates in this single run.

\paragraph{Attribute--behavior relationships.}
Neither of the two tested attribute--behavior associations is statistically significant in the agent. In the human data, greater temporal demand is significantly associated with less pointer--keyboard switching ($\beta=-0.135$), whereas the agent estimate has the opposite direction ($\beta=+0.078$). Greater human effort is significantly associated with more decisions per scored unit ($\beta=+0.110$); the agent estimate has the same direction ($\beta=+0.137$), but is not statistically significant. The human mental-demand association is not statistically significant, so it does not meet the criterion for evaluation in the agent. These findings suggest that capturing how attributes covary is more successful here than reproducing their associations with interaction behavior. Appendix~\ref{app:sense-behavior} provides all coefficients and intervals (Table~\ref{tab:sense-behavior}), sensitivity analyses, and the exploratory hypothesis-selection timing.

\section{Discussion and Future Work}
\label{sec:discussion}

\paragraph{Cognitive models of interaction.}
COGENT represents a simulated user through executable assumptions about knowledge, interpretation, state, and action. A researcher can change what the agent remembers, inspect its understanding of a display, or replace a perceptual or motor process while retaining the surrounding interfaces. The atomic demonstrations show concrete instances of these operations. For cognitive science, this supports a workflow in which a proposed mechanism is implemented, its behavioral consequences are traced, and predictions are compared with human observations. Interface studies can use the same structure to examine how a design interacts with different knowledge or attribute configurations.

The evaluations address complementary levels of correspondence. Next-action prediction holds observed states fixed and tests human choices; executed trajectories include the consequences of the agent's own actions; attribute analyses examine internal relationships and their connection to behavior. Task outcomes, action distributions, and sequence correspondence each describe part of interaction. The 43-task A11y comparison includes unsuccessful attempts, while the shared-success analysis targets behavior conditional on all agents succeeding. This distinction makes cohort choice part of the scientific question.

\paragraph{Calibrating internal dynamics.}
Calibration needs to address relationship ordering, magnitudes, and associations with behavior. Shared linguistic meanings among attribute definitions and dynamics induced by task experience can both contribute to state relationships. Held-out tasks and participants, simpler update baselines, and multiple independent simulator runs would help distinguish these contributions. Future studies can compare predicted trajectories of frustration, effort, or attentiveness with repeated human reports and test whether the corresponding changes in action emerge under matched conditions. SENSE-42's neurophysiological measurements provide a further direction for explicit predictions connecting modeled processes to neural observations.

\paragraph{Module edits and controlling knowledge.}
The module edits in Appendix~\ref{sec:ablation} apply the same operation as the demonstrations in Section~\ref{sec:atomic} across a full task set. Under the default configuration, withholding previous-action assessment lowers exact-action agreement ($p_H=.010$) without changing the pooled action distribution ($p_H=1.000$), restricting candidates to interface affordances lowers agreement mainly through invalid predictions, and full-screen visual access, step-local memory, and objective-level control produce no detectable change. Follow-up edits can separate assessment from consolidation, examine prediction failures under restricted candidate sources, and activate memory and appraisal through calibration. Modularity can be examined by combining independently specified memories, traits, and process implementations in configurations held out from development.

Pretrained knowledge also influences interpretation when the explicit memory store is empty. Unfamiliar interfaces and separate semantic and procedural interventions would help distinguish declared memory from latent familiarity. Specialized perception, retrieval, appraisal, and motor models could replace LLM-backed operations when there is a specific empirical target, such as a visual-search distribution or movement-time curve. Controlled comparisons across model-call budgets would further clarify the contribution of process organization.

\paragraph{Population coverage and use.}
The present evaluations use one model family, selected desktop tasks, and available human reference populations. A11y fidelity targets successful sighted-user interaction; the color-perception demonstration changes information access within the simulator. Extending human reference groups would support comparisons across populations and accessibility needs. Repeated runs, preregistered human studies, and sensitivity analyses across models and prompts can establish when simulated responses preserve the direction and magnitude of an interface effect. The architecture supplies a common structure for these comparisons and for locating the processes associated with a behavioral discrepancy.

\label{main:end}
\clearpage
\subsection*{Reproducibility statement}
The analysis materials include code for action normalization, metric computation, statistical comparisons, and figure generation, together with task selections, model configurations, and machine-readable results. Appendix~\ref{app:provenance} describes the data and reproduction materials.

\subsection*{Ethics statement}
The human comparisons are secondary analyses of existing datasets. The demonstrations use controlled interfaces with fictional request details. The A11y comparison uses sighted-user references, and the color-perception demonstration examines restricted information access within the simulator. Reuse and release of human traces remain subject to the source datasets' consent and access conditions.

\subsection*{AI use statement}
Generative AI was used to help organize the manuscript and improve its language. It also assisted with inspecting analysis code and traces, developing the A11y and AgentNetBench module-edit analysis wrappers, and creating figure-generation and validation scripts. The simulated agents also use LLMs as described in the methods.

\pagebreak
\bibliography{iclr2027_conference}
\bibliographystyle{iclr2027_conference}
\clearpage
\appendix
\section{Architecture and Module Interfaces}
\label{app:design-details}

\subsection{Cognitive motivations and processing operations}
\label{app:design-processes}

Two cognitive theories inspired the architecture: ACT-R and Norman's action cycle. ACT-R explains cognition through specialized perceptual, memory, goal, and motor systems whose coordination depends on information made available through buffers \citep{anderson2004integrated}. This motivates separating the processes that notice, remember, interpret, choose, and execute, while giving them explicit representations through which to interact. Norman's account of action distinguishes forming an intention, specifying and executing an action, and then perceiving, interpreting, and evaluating its consequences \citep{norman1986cognitive}. We use this cycle to organize \cogent's five modules: Observation, Cognition, Control, Action Reasoning, and Execution (Figure~\ref{fig:architecture}). A cycle proceeds from observing the interface toward selecting and executing one physical input, unless the modeled user judges the task complete, pauses, or cannot produce an executable action. The next observation supplies evidence about an executed input's consequences.

The modules coordinate through three shared stores. \emph{Memory} contains long-term facts, procedures, and experiential summaries, together with a bounded working memory of active information and recent experience. Separating stored knowledge from currently available information follows cognitive accounts of memory access, while bounding active contents is motivated by working-memory capacity limits \citep{anderson2004integrated,cowan2001magical}. The \emph{situation model} represents the user's current interpretation of what is happening. We adapt this construct from research on comprehension and memory \citep{zwaan1998situation} to distinguish an interpreted interface from its observable display. The \emph{user-attribute store} contains stable traits and dynamic states, each specified by a meaning and value. Traits condition processing across interactions, whereas states can change with experience. These stores provide continuity between cycles and expose distinct targets for intervention.

\paragraph{Observation and Cognition: from a display to an interpretation.}
Norman's distinction between perceiving a system state and interpreting it motivates separate representations of the available display, noticed information, and understood situation \citep{norman1986cognitive}. Observation captures a screenshot of the current interface. Within Cognition, Perception identifies the regions and cues the modeled user notices and produces a subjective visual workspace. Retrieval uses this evidence and the user's context to activate items from long-term memory. Comprehension integrates the percept, recalled knowledge, and recent experience into the situation model, including an interpretation of the previous action's effect. This separation allows the same visible interface to produce different understandings depending on what the user notices and remembers. It also allows an interpretation to remain incomplete or mistaken and to be revised after further interaction.

\paragraph{Control: from an interpretation to an intention.}
Evaluating a situation and deciding what to pursue require more than recognizing the visible interface. Following Norman's distinction between interpretation and evaluation, Monitor assesses whether the session objective has been achieved from the user's subjective evidence \citep{norman1986cognitive}. Appraise then updates dynamic states in response to that experience, drawing on appraisal accounts in which an event's significance for the individual shapes affective responses \citep{scherer2005emotions}. Executive uses the situation, monitoring result, and updated attributes to maintain or revise an intermediate intention, or to rest, request help, or give up. The maintained intention provides context for subsequent action selection, consistent with accounts of cognitive control that emphasize the active maintenance of goals \citep{miller2001integrative}. Separating these operations gives frustration a role in persistence, while Action Reasoning selects the physical input.

\paragraph{Action Reasoning: from an intention to an operation.}
Human interaction can draw on remembered procedures, information available in the display, and exploratory action. The distinction between knowledge in memory and knowledge available in the environment is central to accounts of interactive behavior \citep{gray2004soft,payne1991display}; exploratory learning supplies another route when a procedure is unfamiliar \citep{rieman1996field}. Motivated by these accounts, Action Reasoning generates candidates from three sources: recalled executable procedures, noticed interface controls, and tentative probes supported by the current situation. Each candidate retains its source, supporting evidence, and expected outcome. Selection commits to one atomic motor intention, such as clicking a particular control or entering a specified string. This makes the basis of a choice inspectable and allows memory-based and display-based strategies to coexist within the same architecture.

\paragraph{Execution: from an operation to feedback.}
Execution realizes the selected intention as a mouse or keyboard input. Separating selection from realization follows both Norman's action cycle and ACT-R's distinction between central and perceptual-motor processing \citep{norman1986cognitive,anderson2004integrated}. A grounding component locates the selected referent, after which the motor component prepares and dispatches the input. The motor implementation can be replaced or perturbed, providing a place to instantiate hypotheses about movement generation and variability \citep{flash1985coordination,vanbeers2004execution}. The subsequent observation closes the cycle: Comprehension interprets what changed, and Control evaluates its significance. An executed input is thus recorded separately from its perceived outcome and from the judgment that the task is complete.

These theories motivate the functional decomposition and the representations passed between modules. The particular grouping into five modules, the serialized processing schedule, and the use of LLMs within several processes are implementation choices. \cogent realizes this functional decomposition through typed records and LLM-backed operations, making its processing assumptions available for testing and revision.

\subsection{Information boundaries}
\label{app:design-boundaries}

Information boundaries preserve the dependencies between processing stages: action selection draws on perceived controls, and execution realizes the selected intention. Each process receives a declared interface containing outputs from preceding stages and the memory, attributes, or task context specified for its role. Access to shared stores follows the same interfaces, providing continuity across cycles. This design takes inspiration from the mediated exchange of information in ACT-R \citep{anderson2004integrated} and makes the effects of an intervention traceable through the processing sequence.

\paragraph{Perceptual and memory access.}
Observation captures a single immutable frame for the cycle. Perception receives that image together with task context, working memory, and relevant attributes, and returns noticed regions, state cues, and a masked subjective image. Retrieval receives the resulting cues and the modeled memory store; returned recalls must identify existing items, and an empty store yields no recall. Comprehension receives the subjective image and records, recalled items, working memory, and the previous action record. These inputs define the evidence available for constructing the situation model, so the visible options for later cognitive processing depend on the controls identified by Perception. Preserving the identities of recalled items also allows researchers to distinguish an activated memory record from newly inferred content.

\paragraph{Evaluation and intention.}
Control operates on the situation and experience established by Cognition. Monitor assesses completion by comparing subjective evidence with the session objective. Appraise receives the interpreted experience, monitoring result, and declared attributes. It updates dynamic states while treating traits as fixed. Executive receives the resulting state and situation, along with permitted task context and working memory, and chooses an intention or a pause. These separate responsibilities distinguish an interpretation of an event, its effect on the user, and the decision about what to pursue next.

\paragraph{Action selection and physical execution.}
Action Reasoning receives subjective perceptual evidence, the situation, current intention, working memory, recalled procedures, and relevant attributes. Pointer candidates must refer to currently noticed actionable regions, and executable memory procedures retain their stored operation templates. Execution receives the selected atomic intention. Its grounding component receives the original screenshot and the selected referent, with its role limited to locating that referent for the fixed operation. Programmatic checks reject changes to the fixed operation and payload before dispatch. This restricted access supports accurate physical grounding while preserving the cognitive decision about what to do. The next observation supplies evidence about the result.

These interfaces specify the evidence supplied to each process and make departures from the intended processing path easier to identify. LLM parameters also supply pretrained knowledge: a model may recognize an icon or infer a procedure when the explicit memory store is empty. Section~\ref{sec:discussion} considers experiments for separating the contributions of supplied information and pretrained knowledge.

\section{Intervenability Demonstration Evidence}
\label{app:demos}

\subsection{Interventions and recorded evidence}
\label{app:demos-scope}

Each demonstration comprises one pair selected during development and illustrates an intervention and its recorded consequences. Evidence includes the configuration, initial and subsequent screenshots, executed inputs, intermediate module outputs, and task outcomes. Table~\ref{tab:demo-boundaries} specifies the changed component and the supported interpretation for each pair; panel letters follow Figure~\ref{fig:atomic}.

\begin{table}[htbp]
\centering
\caption{Intervention and evidence boundaries for the six paired demonstrations.}
\label{tab:demo-boundaries}
\small
\begin{tabularx}{\linewidth}{@{}lXX@{}}
\toprule
Demo & Changed component & Supported contrast \\
\midrule
(a) Icon memory & Empty versus nine-item calibration memory & Six executed inputs versus one; semantic and procedural knowledge are changed together. \\
(b) Close window & Executable Alt+F4 memory & Native close click versus keyboard chord; one input and successful closure in both. \\
(c) Carefulness & Stable trait, low versus high & Direct submission versus review then confirmation; trait values remain fixed. \\
(d) Frustration & Initial dynamic-state value & Two attempts and completion versus one attempt then Executive help-seeking pause. \\
(e) Motor & Direct versus minimum-jerk execution & Different measured movement time and intermediate positions; common button target, slightly different endpoints. \\
(f) Perception & Image transformation at the perceptual boundary & Direct Ready selection versus revealing labels first; raw starting interface unchanged. \\
\bottomrule
\end{tabularx}
\end{table}

\subsection{Memory walkthrough and execution evidence}
\label{app:demos-memory}

\paragraph{Demo (a): icon and procedural knowledge.}
This demonstration shows how supplied semantic and procedural knowledge changes recognition and action selection. Two agents receive the same unfamiliar dock and the goal of opening File Manager. With empty long-term memory, the agent retrieves nothing and selects Document three times, Browser once, Document again, and finally File Manager. With the full nine-item calibration memory, it retrieves dock knowledge and an executable single-click procedure, then opens File Manager on its first action (Figure~\ref{fig:atomic}a). The memory-sourced candidate explicitly cites the recalled procedure and the noticed target. The empty-memory versus full-calibration-memory intervention changes semantic and procedural knowledge together, connecting recalled knowledge to the selected action.

The full calibration memory contains eight semantic items and one procedure. Relevant semantic item M4 maps the dock identities, while procedure M5 instructs a single click on the blue smiling Finder face and carries an executable motor template. The memory condition's first selected candidate cites M5 and perceptual region P1. Empty memory produces no recalled items. Both agents initially notice the blue face; the recalled record supplies its identity and the procedure for opening it.

The no-memory action sequence provides a useful check on the interpretation of model prose. In its fourth step, the candidate description names the smiling-face icon, but the dispatched coordinate lands in the Browser hit box. The fixture records Browser. The sixth dispatched coordinate falls in File Manager and produces success. Accordingly, figure labels and action counts follow executed inputs and observed outcomes. The cognitive and execution traces jointly reveal the route from candidate selection through grounding to the interface response.

\paragraph{Demo (b): alternative successful actions.}
This demonstration shows how remembering an executable procedure changes the strategy used to achieve the same outcome. Both agents close an identical browser window in one physical input. The empty-memory agent selects the visible native close control. With an executable Alt+F4 memory, the agent selects the recalled keyboard operation (Figure~\ref{fig:atomic}b). The two successful strategies use different candidate sources and action forms.

\begin{figure}[htbp]
\centering
\includegraphics[width=\linewidth]{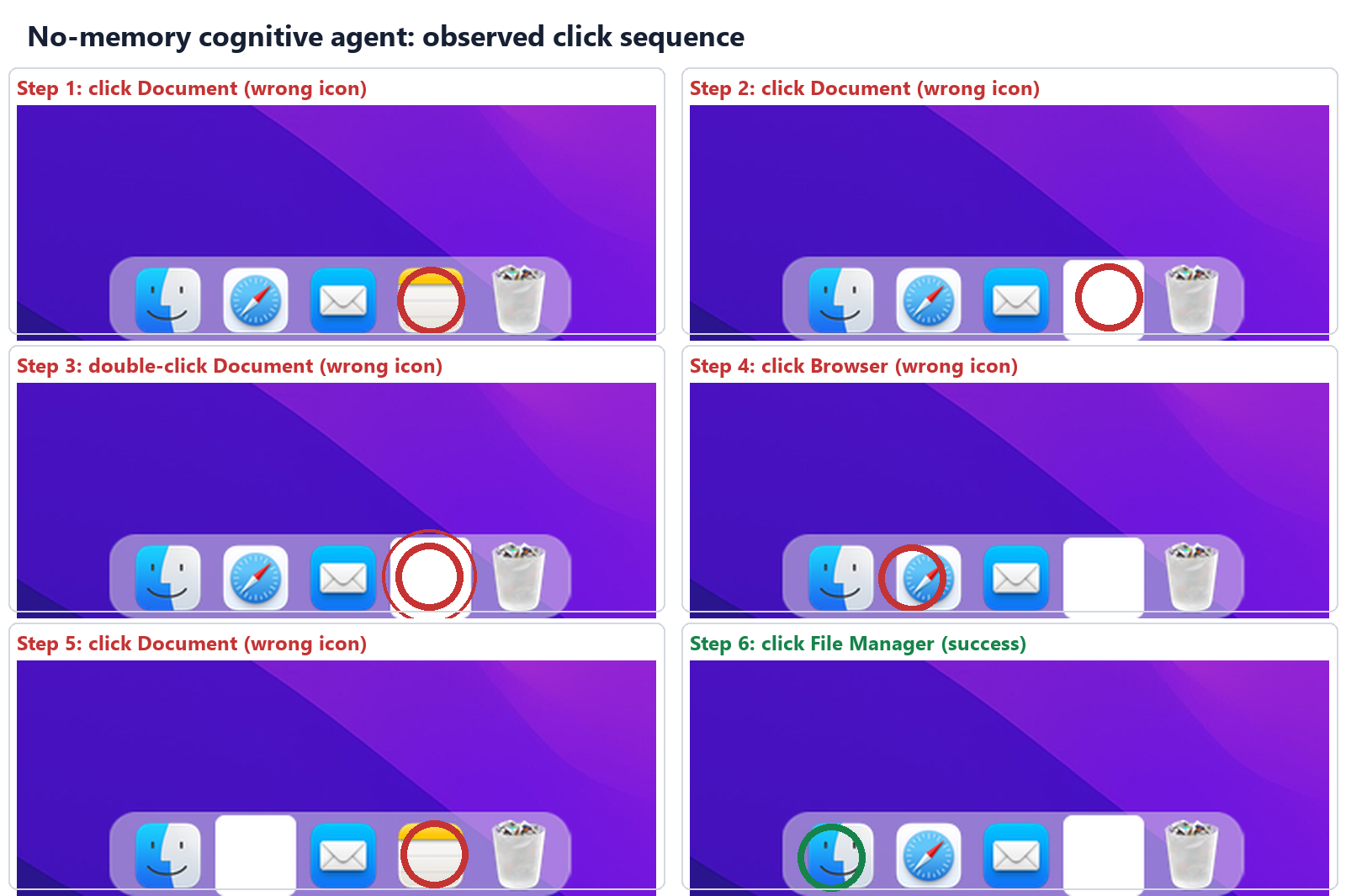}
\caption{Complete empty-memory icon walkthrough. The executed selections lead to File Manager on the sixth action. The screenshots show the consequences of each input; the main figure compares the empty and full calibration-memory conditions.}
\label{fig:icon-appendix}
\end{figure}

\subsection{Attributes and control}
\label{app:demos-attributes}

\paragraph{Demo (c): carefulness.}
This demonstration shows how carefulness, a stable user attribute, changes whether the agent reviews its work before submission. The same prepared appointment request can be submitted immediately or reviewed first. Low carefulness yields direct submission; high carefulness yields a review action followed by confirmation. Both conditions succeed, using one and two physical inputs respectively (Figure~\ref{fig:atomic}c). Carefulness is specified as a general tendency toward checking and restraint. The pair illustrates how this attribute changes the route through an interface while the task objective remains fixed.

\paragraph{Demo (d): frustration.}
This demonstration shows how the dynamic state of frustration changes persistence after an unsuccessful attempt. A report is initially unavailable. Under low initial frustration, the agent attempts access twice and completes the task. Under high initial frustration, it attempts access once and then the Executive pauses to request help (Figure~\ref{fig:atomic}d). The \texttt{ask\_help} pause is an internal control decision recorded by the Executive. The pair shows frustration influencing persistence through the intention-control pathway.

\subsection{Motor and perceptual measurements}
\label{app:demos-processes}

\paragraph{Demo (e): motor execution.}
This demonstration shows how replacing the motor process changes the movement used to execute a fixed action intention. The selected referent is the same Continue button, while the motor component uses direct positioning or a minimum-jerk trajectory. Independent cursor observations show movement intervals of approximately 0.0077 and 0.5606 seconds, with zero and 35 distinct intermediate positions respectively; both clicks land inside the intended button (Figure~\ref{fig:atomic}e). Target identity is held fixed, with slightly different endpoints within the button. The plot follows the observed coordinates, preserving the measured movement profile. Smooth-movement theory motivates the substitution \citep{flash1985coordination}.

\paragraph{Demo (f): color perception.}
This demonstration shows how changing color perception changes the strategy used to identify a target. Both agents must open the Ready item on a page where green indicates Ready and red indicates Unavailable. We compare two Perception configurations: \emph{Condition 1} receives the original screenshot, while \emph{Condition 2} applies a deuteranopia image transform that makes the two status colors difficult to distinguish. Under Condition 1, the agent identifies the green Ready item and opens it directly with one click. Under Condition 2, the agent first clicks \emph{Show status labels}, then uses the revealed text to identify and open the Ready item with a second click. Both conditions succeed (Figure~\ref{fig:atomic}f). The task, raw interface, memory, attributes, and downstream architecture remain fixed. The pair illustrates how altered perceptual evidence leads to an additional information-seeking action before target selection.

The cursor measurements use the independent native observation stream. Movement begins at the last sample before coordinates first change and ends at the last changed coordinate. For a point $p(t)$, start $p_0$, and endpoint $p_1$, plotted progress is
\begin{equation}
u(t)=\frac{(p(t)-p_0)^\top(p_1-p_0)}{\|p_1-p_0\|^2}.
\label{eq:motor-progress}
\end{equation}
We retain all observed samples between those boundaries and connect them directly. The direct movement spans one observation interval of 0.0077 seconds, placing its duration at the sampling resolution. The minimum-jerk movement spans approximately 0.5606 seconds, with 35 distinct sampled positions excluding the start and endpoint. The commanded duration is 0.6 seconds; commanded and observed durations are reported as different quantities. In Figure~\ref{fig:atomic}e, dots mark observed samples, and the inset overlays the direct (circle) and minimum-jerk (plus) click endpoints on the recorded Continue-button crop.

The perception figure uses the actual model-input images from the same initial frame, with matched crops of the two status colors.

\section{AgentNetBench Protocol and Supplementary Results}
\label{app:agentnet}

We evaluate 100 tasks and 1,208 screenshot decisions from OpenCUA/AgentNet \citep{wang2025opencuaopenfoundationscomputeruse}. At each decision, an agent predicts the next human action from the current screenshot, task, and preceding context. The recorded human action determines the next state, holding the observation sequence fixed across agents. Direct baselines retain screenshot/action context, whereas \cogent retains its cognitive state; \cogent and \cuauser additionally receive the same fixed participant-context description.

Exact-action matching checks the normalized action structure and relevant keys or text; pointer actions must fall within an accepted target region. Action-type accuracy uses the 1,111 decisions whose human action maps to the shared eight-class taxonomy. A secondary subset contains 82 decisions across 53 tasks with annotated alternative actions. It distinguishes agreement with any annotated-valid action from agreement with the recorded human choice. The alternatives form a curated set of workable actions. We estimate uncertainty using 10,000 paired task-cluster bootstrap samples. Comparisons use paired-difference intervals; the 95\% intervals are marginal and unadjusted for multiple comparisons.

\begin{table}[htbp]
\centering
\caption{\textbf{AgentNetBench: full prediction results.} Percentages with 95\% task-bootstrap intervals. Exact/type denominators are 1,208/1,111. The alternative-action subset has 82 decisions across 53 tasks; human choice conditions on 50, 50, and 47 valid predictions. Bold marks the highest estimate, including ties.}
\label{tab:agentnet-full}
\small\begin{tabular}{@{}lcccc@{}}
\toprule
Agent & Exact action $\uparrow$ & Action type $\uparrow$ & Any valid $\uparrow$ & Human choice $\uparrow$ \\
\midrule
COGENT & \textbf{59.8} & \textbf{84.1} & \textbf{61.0} & \textbf{32.0} \\
 & [56.4, 63.0] & [81.6, 86.5] & [48.7, 72.3] & [18.0, 47.6] \\
CUA & 52.2 & 79.0 & \textbf{61.0} & 18.0 \\
 & [48.8, 55.3] & [76.1, 81.9] & [50.0, 71.6] & [7.5, 29.2] \\
CUA-User & 53.7 & 80.9 & 57.3 & 29.8 \\
 & [50.3, 57.1] & [78.2, 83.6] & [46.9, 67.0] & [17.4, 43.2] \\
\bottomrule
\end{tabular}

\end{table}

In the alternative-action subset, any-annotated-valid accuracy is 61.0\% for both \cogent and \cua and 57.3\% for \cuauser. The paired differences are $0.00$ points (95\% CI [$-10.13,+9.88$]) and $+3.66$ points [$-6.85,+14.12$], respectively. Conditional agreement with the recorded human choice is 32.0\%, 18.0\%, and 29.8\%; the corresponding \cogent--baseline differences are $+14.00$ points [$-0.45,+30.02$] and $+2.21$ points [$-9.22,+14.47$]. All four intervals span zero, leaving these differences inconclusive. The two measures distinguish agreement with the recorded human choice from agreement with any annotated workable action.

\section{Module Edits Across Tasks}
\label{sec:ablation}

The intervenability demonstrations show how edits to a simulated user change behavior in individual cases. Here, we examine how five edits to processing modules affect human-action prediction across an entire task set. Each edit implements a different assumption about the simulated user: what visual information is available, what memory persists, how intentions are maintained, which actions are considered, or how previous actions are evaluated. This analysis extends the demonstrations from individual interaction traces to aggregate behavioral fidelity.

\subsection{Evaluation setting}

We compare the five edited configurations with full \cogent on the same 100 AgentNetBench tasks and 1,208 recorded human-action targets, using five replicates per condition. All configurations use GPT-5.4, original-size screenshots, at most four action candidates, and no motor perturbation. Each prediction receives the current screenshot and up to two preceding screenshot--action pairs. Recorded human actions determine subsequent interface states, allowing us to compare the edits under the same observed task history. The full configuration begins with empty authored long-term memory and user-attribute stores. Retrieval and appraisal are inactive in this setting, and the prediction protocol does not run consolidation.

We measure exact-action agreement, action-type accuracy, and pooled JS distance from the human action-type distribution. Exact accuracy uses all 6,040 decisions per condition, including 485 human termination targets. Action-type accuracy uses the 5,555 targets with a classified physical human action. Invalid predictions, including adapter errors, remain in the accuracy denominators and count as incorrect. Accuracy estimates pool decisions across tasks and replicates, giving tasks with more targets greater weight. JS compares pooled classified physical-action counts; predicted termination and unclassifiable outputs are excluded, so the agent distribution denominators vary by condition.

The comparison uses the full configuration from this experiment, whose exact and action-type accuracies are 57.32\% and 81.21\%, respectively. Uncertainty is estimated with 2,000 paired hierarchical bootstrap resamples of replicates and tasks. The same sampled clusters are used for both conditions in each contrast. Confidence intervals are marginal 95\% intervals; significance on the primary measures is assessed using Holm-adjusted $p_H$ across the ten accuracy contrasts. The five exploratory JS contrasts form a separate Holm family. Table~\ref{tab:ablation-main} summarizes the estimates and adjusted comparisons.

\begin{table}[htbp]
\centering
\caption{\textbf{Effects of module edits on AgentNetBench.} Parentheses give Holm-adjusted $p_H$ versus full \cogent; bold marks significant accuracy reductions ($p_H<.05$). Exact/type denominators are 6,040/5,555. JS compares pooled classified physical actions.}
\label{tab:ablation-main}
\small\setlength{\tabcolsep}{4pt}
\begin{tabular}{@{}lrrr@{}}
\toprule
Condition & Exact (\%) $\uparrow$ & Type (\%) $\uparrow$ & JS $\downarrow$ \\
\midrule
Full \cogent & 57.32 & 81.21 & 0.0561 \\
Raw downstream image & 58.26 (0.978) & 82.00 (1.000) & 0.0502 (1.000) \\
Memoryless & 56.87 (1.000) & 81.40 (1.000) & 0.0553 (1.000) \\
Objective-only control & 58.73 (0.161) & 82.02 (1.000) & 0.0573 (1.000) \\
Interface-only action & \textbf{53.11} (0.010) & \textbf{75.84} (0.010) & 0.0490 (1.000) \\
No lagged feedback & \textbf{53.63} (0.010) & 80.59 (1.000) & 0.0561 (1.000) \\
\bottomrule
\end{tabular}
\end{table}

\subsection{Modeling scenarios and their effects}

\paragraph{Perception: make the entire screen available to reasoning.}
The \emph{raw downstream image} edit implements an idealized full-access assumption: visual information from every screen region is available to downstream cognition. It replaces Perception's attention-masked image with the original screenshot while retaining the structured percept and noticed-region records. This allows reasoning to use visual evidence beyond the regions selected by the default perceptual process. Exact-action accuracy is 58.26\% and action-type accuracy is 82.00\%; neither differs significantly from full \cogent ($p_H=.978$ and $1.000$). The JS difference is also not statistically significant ($p_H=1.000$). Thus, this full-access edit produces no statistically significant change in the evaluated fidelity measures.

\paragraph{Memory processing: reason without carried working-memory records.}
The \emph{memoryless} edit models a user whose working-memory contents do not persist between processing cycles and whose stored knowledge does not supply recalled procedures. It clears working-memory records before each cognitive cycle, disables retrieval and memory-based action proposals, and removes the preference for memory candidates. The supplied recent screenshot--action history remains available. Because authored long-term memory is empty in the full configuration, the active contrast primarily concerns carried working-memory contents. Exact-action accuracy is 56.87\% and action-type accuracy is 81.40\%; neither accuracy difference nor the JS difference is statistically significant (all $p_H=1.000$).

\paragraph{Control: maintain a fixed task-level intention.}
The \emph{objective-only control} edit models a user who pursues the stated task objective without forming intermediate intentions. The Executive maintains the objective itself as the current intention and does not select meta-actions such as requesting help; Appraisal is frozen. With no configured dynamic attributes, the active change concerns executive control. Exact-action accuracy is 58.73\% and action-type accuracy is 82.02\%. Although both point estimates exceed the full configuration's, the differences are not statistically significant after correction ($p_H=.161$ and $1.000$), and neither is the JS difference ($p_H=1.000$). The results therefore provide no significant improvement on the primary accuracy measures. A secondary analysis finds higher annotated-valid action accuracy, at 61.54\% versus 59.87\%, a gain of 1.67 percentage points (exploratory unadjusted $p=.009$). This measure credits any annotated workable action, including alternatives to the recorded human choice.

\paragraph{Action Reasoning: consider only interface-grounded candidates.}
The \emph{interface-only action} edit models a user whose action choices are generated from recognized interface affordances, without separate memory-based or exploratory proposals. It restricts Action Reasoning to the interface candidate source. Since the full configuration generates no memory candidates in this experiment, the operative restriction removes exploratory proposals. Exact-action accuracy falls to 53.11\% and action-type accuracy to 75.84\%, significant reductions of 4.21 and 5.36 percentage points (both $p_H=.010$). JS distance does not change significantly ($p_H=1.000$). The invalid-output rate increases by 7.24 points (exploratory unadjusted $p=.001$). On the 5,468 decisions where neither condition has an adapter error, the exact-accuracy difference is $+0.11$ percentage points and is not statistically significant (Table~\ref{tab:module-edit-sensitivity}). This pattern links the aggregate accuracy loss to the reliability of producing a valid prediction when candidate sources are restricted.

\paragraph{Comprehension: interpret the current screen without assessing the preceding action.}
The \emph{no lagged feedback} edit models a user who interprets the current interface without explicitly evaluating whether the previous action produced its expected outcome. It withholds prior-action records from Comprehension's assessment process and suppresses the resulting outcome assessment and experience records. Current-screen interpretation and action history for other processes remain available. Exact-action accuracy falls to 53.63\%, a significant reduction of 3.69 points ($p_H=.010$); action-type accuracy and JS distance do not change significantly (both $p_H=1.000$). On the 5,905 decisions where neither condition has an adapter error, exact accuracy remains 3.88 percentage points lower, a statistically significant reduction under the exploratory, unadjusted comparison (Table~\ref{tab:module-edit-sensitivity}). Of the 223 lost exact matches, 106 concern human stopping decisions, illustrating how explicit outcome assessment can affect when the simulated user stops as well as which physical action is predicted.

\subsection{Complete comparisons and prediction reliability}

Tables~\ref{tab:module-edit-accuracy} and~\ref{tab:module-edit-distribution} report every primary accuracy and exploratory JS contrast. Objective-only control's marginal exact-accuracy interval excludes zero, but its Holm-adjusted $p_H=.161$ does not meet the significance threshold. All five JS contrasts are statistically nonsignificant.

\begin{table}[htbp]
\centering
\caption{\textbf{Primary accuracy contrasts for module edits.} Accuracies are percentages; $\Delta$ is edited minus full accuracy in percentage points. CIs are marginal 95\% intervals; bold marks $p_H<.05$ after Holm adjustment across the ten contrasts.}
\label{tab:module-edit-accuracy}
\footnotesize\setlength{\tabcolsep}{4pt}
\begin{tabular}{@{}llrrcr@{}}
\toprule
Condition & Metric & Accuracy & $\Delta$ & 95\% CI & $p_H$ \\
\midrule
Raw downstream image & Exact & 58.26 & $+0.94$ & $[-0.33, 2.24]$ & 0.978 \\
 & Type & 82.00 & $+0.79$ & $[-0.73, 2.11]$ & 1.000 \\
Memoryless & Exact & 56.87 & $-0.45$ & $[-1.83, 0.88]$ & 1.000 \\
 & Type & 81.40 & $+0.20$ & $[-1.07, 1.31]$ & 1.000 \\
Objective-only control & Exact & 58.73 & $+1.41$ & $[0.21, 2.62]$ & 0.161 \\
 & Type & 82.02 & $+0.81$ & $[-0.49, 2.16]$ & 1.000 \\
Interface-only action & Exact & 53.11 & $-4.21$ & $[-6.20, -2.33]$ & \textbf{0.010} \\
 & Type & 75.84 & $-5.36$ & $[-8.05, -2.71]$ & \textbf{0.010} \\
No lagged feedback & Exact & 53.63 & $-3.69$ & $[-4.94, -2.42]$ & \textbf{0.010} \\
 & Type & 80.59 & $-0.61$ & $[-2.09, 0.91]$ & 1.000 \\
\bottomrule
\end{tabular}
\end{table}

\begin{table}[htbp]
\centering
\caption{\textbf{Pooled action-distribution contrasts for module edits.} $\Delta$ is edited minus full JS distance; positive values indicate greater distance from humans. Full JS is 0.05613. CIs are marginal 95\% intervals; $p_H$ adjusts the five exploratory contrasts.}
\label{tab:module-edit-distribution}
\footnotesize\setlength{\tabcolsep}{4pt}
\begin{tabular}{@{}lrrcr@{}}
\toprule
Condition & JS & $\Delta$ JS & 95\% CI & $p_H$ \\
\midrule
Raw downstream image & 0.05016 & $-0.00597$ & $[-0.01673, 0.00527]$ & 1.000 \\
Memoryless & 0.05531 & $-0.00082$ & $[-0.01525, 0.01356]$ & 1.000 \\
Objective-only control & 0.05732 & $+0.00119$ & $[-0.01201, 0.01603]$ & 1.000 \\
Interface-only action & 0.04901 & $-0.00712$ & $[-0.02060, 0.00782]$ & 1.000 \\
No lagged feedback & 0.05612 & $-0.00001$ & $[-0.01503, 0.01385]$ & 1.000 \\
\bottomrule
\end{tabular}
\end{table}

Table~\ref{tab:module-edit-sensitivity} compares predictions for which neither condition has an adapter error. Other invalid predictions and termination outputs remain included. These post-hoc subsets differ across edits, and their intervals are unadjusted. The interface-only exact-accuracy difference is not significant within its matched subset, whereas the no-feedback deficit persists. This comparison helps distinguish the contribution of prediction reliability from fidelity differences that remain among predictions produced without adapter errors.

\begin{table}[htbp]
\centering
\caption{\textbf{Sensitivity to adapter errors.} Changes are edited minus full accuracy in percentage points, conditional on neither prediction having an adapter error. $N_E$ and $N_T$ are the exact- and action-type denominators. Brackets give unadjusted 95\% CIs.}
\label{tab:module-edit-sensitivity}
\footnotesize\setlength{\tabcolsep}{3pt}
\begin{tabular}{@{}lrcrc@{}}
\toprule
Condition & $N_E$ & Exact change [95\% CI] & $N_T$ & Type change [95\% CI] \\
\midrule
Raw downstream image & 5934 & $+0.54$ $[-0.69, 1.83]$ & 5463 & $+0.40$ $[-0.54, 1.28]$ \\
Memoryless & 5911 & $-0.69$ $[-1.75, 0.36]$ & 5444 & $+0.04$ $[-0.91, 0.90]$ \\
Objective-only control & 5910 & $+1.18$ $[0.02, 2.28]$ & 5443 & $+0.61$ $[-0.33, 1.64]$ \\
Interface-only action & 5468 & $+0.11$ $[-1.04, 1.31]$ & 5060 & $+0.40$ $[-0.82, 1.68]$ \\
No lagged feedback & 5905 & $-3.88$ $[-4.99, -2.77]$ & 5439 & $-0.72$ $[-1.78, 0.32]$ \\
\bottomrule
\end{tabular}
\end{table}

\subsection{Implications for simulation}

These edits specify alternative processing assumptions whose consequences can be examined across tasks. Restricting action candidates and removing explicit previous-action assessment significantly reduce exact-action agreement in this configuration, whereas full-screen visual access, removal of carried memory, and objective-level control yield no statistically significant changes on the primary accuracy measures after correction. None of the five edits significantly changes pooled action-type JS distance. The contrast between these measures shows why evaluating edits requires both action-level agreement and aggregate distributions: changes in when an action occurs, whether a valid prediction is produced, or when a user stops can coexist with similar overall action frequencies.

The findings apply to a bounded-history configuration with empty authored memory and attributes. Calibration memory, active dynamic states, and longer interaction histories provide further settings in which to examine the same edits. The five replicates capture repeated model predictions over the same task corpus; additional models and task domains would broaden the settings in which these processing assumptions can be examined.

\section{A11y Cohorts, Success, and Sensitivity Analyses}
\label{app:a11y}

\subsection{Fidelity metrics}
\label{app:fidelity-metrics}

The eight action classes are click, double click, right click, drag, scroll, text entry, single key, and key chord. For action-distribution comparison, let $p$ and $q$ be the normalized pooled agent and human counts on the selected tasks. Using base-2 logarithms, Jensen--Shannon distance is
\begin{equation}
d_{\rm JS}(p,q)=\sqrt{\tfrac12 D_{\rm KL}(p\Vert m)+\tfrac12 D_{\rm KL}(q\Vert m)},
\qquad m=(p+q)/2.
\label{eq:js}
\end{equation}
Counts are pooled before normalization; longer trajectories therefore contribute more actions. This is the same distribution estimator used for AgentNetBench.

We combine consecutive printable human key events into text-entry episodes and nearby human scroll events into scroll episodes using a 0.75-second gap. We exclude controller activity and noninteraction pointer movement. Agent sequences contain executed physical commands. The taxonomy measures interaction form by grouping actions across target identities and semantics.

For action sequences $x$ and $y$, let $L=\min(|x|,|y|)$. The matched-horizon sequence distance is
\begin{equation}
d_{\rm seq}(x,y)=\operatorname{Lev}(x_{1:L},y_{1:L})/L.
\label{eq:sequence}
\end{equation}
Both-empty and one-empty sequences receive distances zero and one respectively. All own-success agent trajectories are nonempty; the matched cohort includes zero-action errors for \cogent and \cua on task 14. The comparison omits unmatched suffixes, so sequence distance must be read together with trajectory length and outcome. Agent--human distances average eligible human references within each task, and task-level sequence statistics receive equal weight. The human-normalized gap in Equation~\ref{eq:hng} scales this distance by the mean over distinct human pairs on the same task.

Sequence statistics and trajectory lengths average humans within task and then weight tasks equally. Action-type shares and their Jensen--Shannon distance use pooled counts. A11y intervals use 2,000 whole-task bootstrap resamples. Matched comparisons use the same sampled tasks for all systems and report differences of pooled distances or mean sequence statistics. For a distribution estimate, each sampled task contributes all its agent and human action counts, and the pooled distributions and distance are recomputed within that draw. The appendix's separate own-success cohorts are resampled independently. These exploratory intervals are unadjusted for multiple comparisons; they preserve within-task dependence and quantify variation over the evaluated tasks.

\subsection{Task selection and outcome assessment}
\label{app:a11y-archive}

We selected 43 Windows tasks and compare the original three agents (\cogent, \cua, and \cuauser) using GPT-5.4, each with a 30-step interaction budget. Every agent is evaluated on every selected task, yielding 129 agent trajectories. The human reference contains 339 successful sighted-user trajectories on these same tasks.

Task outcomes are determined by independent review of complete interaction traces. An episode is successful when the review confirms the task objective, failed when it confirms that the objective was not achieved, and unverifiable when the evidence does not support either judgment. Unverifiable outcomes are reported separately from verified failures. The original three-agent fidelity comparison includes all 129 episodes, covering successful, failed, and unverifiable outcomes.

Action sequences use chronological mouse and keyboard commands. Wait commands are excluded from normalized action counts, which therefore measure a different unit from reported interaction steps. Episodes were collected across interrupted sessions.

\subsection{Success and failure breakdown}
\label{app:a11y-outcomes}

Table~\ref{tab:success-rates} reports outcomes on the 43 selected tasks. Success rates use the same denominator of 43 tasks for every agent. Verified successes, verified failures, and unverifiable outcomes partition each agent's evaluated trajectories.

\begin{table}[htbp]
\centering
\caption{\textbf{A11y-CUA: outcomes on the 43 selected tasks.} Success rates are verified successes divided by 43. Unverifiable outcomes are distinct from verified failures; all outcomes enter the main fidelity comparison.}
\label{tab:success-rates}
\label{tab:a11y-matched-outcomes}
\small\begin{tabular}{@{}lrrrrr@{}}
\toprule
Agent & Tasks & Success & Failure & Unverif. & Success rate \\
\midrule
COGENT & 43 & 24 & 15 & 4 & 55.8\% \\
CUA & 43 & 20 & 21 & 2 & 46.5\% \\
CUA-User & 43 & 19 & 22 & 2 & 44.2\% \\
\bottomrule
\end{tabular}

\end{table}

\subsection{Matched all-attempts comparison}
\label{app:a11y-matched}

All 43 selected tasks form the original three-agent subset of the matched cohort reported in the main paper (Table~\ref{tab:a11y-main}). This subset contains 129 agent episodes, including failed and unverifiable outcomes; task 14 contributes zero-action errors for \cogent and \cua. Sequence distance is one when only one sequence is empty. Table~\ref{tab:a11y-matched-outcomes} gives the outcome counts. The common human pool contains 10,131 actions; agent pools contain 809, 703, and 707 actions. Mean semantic-action lengths are 18.81 [95\% CI: 15.72, 21.88] for \cogent, 16.35 [12.98, 19.58] for \cua, and 16.44 [12.81, 19.93] for \cuauser, with a task-averaged human mean of 30.09 [24.94, 36.74]. These marginal intervals describe uncertainty in each mean length. The equal-task human--human sequence distance is 0.344. Table~\ref{tab:a11y-matched-contrasts} reports all paired metric contrasts for this cohort.

\begin{table}[htbp]
\centering
\caption{\textbf{A11y-CUA: paired differences across 43 matched tasks.} Negative values favor the first agent. Bold differences have intervals excluding zero. Bounds are nominal, unadjusted 95\% paired task-bootstrap intervals. Both \cogent comparisons exclude zero for pooled JS and raw sequence distance, but include zero for sequence HNG.}
\label{tab:a11y-matched-contrasts}
\small\begin{tabular}{@{}llrc@{}}
\toprule
Contrast & Metric & Difference & 95\% CI \\
\midrule
COGENT $-$ CUA & Pooled JS & \textbf{-0.098} & $[-0.161, -0.041]$ \\
COGENT $-$ CUA & Sequence & \textbf{-0.084} & $[-0.147, -0.020]$ \\
COGENT $-$ CUA & Seq. HNG & -0.177 & $[-0.465, +0.147]$ \\
COGENT $-$ CUA-User & Pooled JS & \textbf{-0.146} & $[-0.219, -0.078]$ \\
COGENT $-$ CUA-User & Sequence & \textbf{-0.101} & $[-0.172, -0.033]$ \\
COGENT $-$ CUA-User & Seq. HNG & -0.133 & $[-0.429, +0.173]$ \\
CUA $-$ CUA-User & Pooled JS & -0.048 & $[-0.112, +0.022]$ \\
CUA $-$ CUA-User & Sequence & -0.017 & $[-0.067, +0.034]$ \\
CUA $-$ CUA-User & Seq. HNG & +0.045 & $[-0.252, +0.339]$ \\
\bottomrule
\end{tabular}

\end{table}

These intervals quantify variation across the evaluated tasks. Double- and right-click counts are zero in the human reference; the comparison retains this difference in action representation. Distribution comparisons pool action counts; sequence comparisons weight tasks equally.

\subsection{Matched shared-success comparison}
\label{app:a11y-shared}

Tasks 7, 10, 21, 22, 24, 25, 26, 27, 28, 29, 30, 31, 32, and 33 are successful for all three agents: 42 agent trajectories and 111 human references. Each agent contributes 14 verified successful trajectories. This analysis targets behavior conditional on joint success.

\begin{table}[htbp]
\centering
\caption{\textbf{A11y-CUA: fidelity on 14 shared-success tasks.} Bold marks the lowest point estimate; paired uncertainty is reported in Table~\ref{tab:a11y-shared-contrasts}. JS pools action counts; sequence statistics weight tasks equally. Brackets are 95\% task-bootstrap intervals.}
\label{tab:a11y-shared}
\small\begin{tabular}{@{}lrccc@{}}
\toprule
Agent & Tasks & JS distance $\downarrow$ & Sequence distance $\downarrow$ & HNG $\downarrow$ \\
\midrule
COGENT & 14 & \textbf{0.346} & \textbf{0.469} & 1.987 \\
 & & [0.306, 0.421] & [0.402, 0.538] & [1.297, 2.920] \\
CUA & 14 & 0.406 & 0.535 & \textbf{1.921} \\
 & & [0.322, 0.533] & [0.425, 0.643] & [1.456, 2.443] \\
CUA-User & 14 & 0.485 & 0.611 & 2.311 \\
 & & [0.396, 0.614] & [0.483, 0.746] & [1.659, 3.030] \\
\bottomrule
\end{tabular}

\end{table}

\begin{table}[htbp]
\centering
\caption{\textbf{A11y-CUA: paired differences on 14 shared-success tasks.} Negative values favor the first agent; bold differences have intervals excluding zero. Intervals are unadjusted 95\% task-bootstrap intervals. \cogent--\cua intervals include zero on all three metrics. Against \cuauser, pooled JS excludes zero and sequence distance only narrowly excludes zero; sequence HNG includes zero.}
\label{tab:a11y-shared-contrasts}
\small\begin{tabular}{@{}llrc@{}}
\toprule
Contrast & Metric & Difference & 95\% CI \\
\midrule
COGENT $-$ CUA & Pooled JS & -0.060 & $[-0.197, +0.064]$ \\
COGENT $-$ CUA & Sequence & -0.066 & $[-0.187, +0.059]$ \\
COGENT $-$ CUA & Seq. HNG & +0.065 & $[-0.542, +0.813]$ \\
COGENT $-$ CUA-User & Pooled JS & \textbf{-0.139} & $[-0.272, -0.024]$ \\
COGENT $-$ CUA-User & Sequence & \textbf{-0.142} & $[-0.285, -0.001]$ \\
COGENT $-$ CUA-User & Seq. HNG & -0.324 & $[-0.880, +0.190]$ \\
CUA $-$ CUA-User & Pooled JS & -0.079 & $[-0.170, +0.038]$ \\
CUA $-$ CUA-User & Sequence & \textbf{-0.076} & $[-0.159, -0.003]$ \\
CUA $-$ CUA-User & Seq. HNG & \textbf{-0.390} & $[-0.890, -0.024]$ \\
\bottomrule
\end{tabular}

\end{table}

\subsection{Own-success cohorts, action counts, and lengths}
\label{app:a11y-own}

For completeness, Table~\ref{tab:a11y-own-success} conditions separately on each agent's verified successes. The differing task sets make these comparisons descriptive. \cogent has lower raw JS and sequence distances, whereas \cua has lower HNG. HNG averages task-specific ratios to human variability, so its ranking reflects both raw distances and the human variability on each selected task.

\begin{table}[htbp]
\centering
\caption{\textbf{A11y-CUA: descriptive fidelity on each agent's successful tasks.} Task sets differ across rows. Lower values indicate closer human correspondence; bold marks the lowest point estimate. Brackets give 95\% task-bootstrap intervals. JS pools action counts; sequence distance and HNG average tasks equally.}
\label{tab:a11y-own-success}
\small\begin{tabular}{@{}lrccc@{}}
\toprule
Agent & Tasks & JS distance $\downarrow$ & Sequence distance $\downarrow$ & HNG $\downarrow$ \\
\midrule
COGENT & 24 & \textbf{0.323} & \textbf{0.489} & 2.084 \\
 & & [0.282, 0.386] & [0.437, 0.537] & [1.486, 2.869] \\
CUA & 20 & 0.382 & 0.553 & \textbf{1.925} \\
 & & [0.334, 0.469] & [0.465, 0.634] & [1.513, 2.393] \\
CUA-User & 19 & 0.448 & 0.616 & 2.219 \\
 & & [0.382, 0.542] & [0.503, 0.727] & [1.741, 2.778] \\
\bottomrule
\end{tabular}

\end{table}

\paragraph{COGENT successful task IDs.} 1, 7, 9, 10, 11, 18, 20, 21, 22, 24, 25, 26, 27, 28, 29, 30, 31, 32, 33, 34, 42, 43, 52, 54.

\paragraph{CUA successful task IDs.} 1, 2, 7, 9, 10, 21, 22, 24, 25, 26, 27, 28, 29, 30, 31, 32, 33, 34, 50, 60.

\paragraph{CUA-User successful task IDs.} 2, 7, 10, 18, 20, 21, 22, 24, 25, 26, 27, 28, 29, 30, 31, 32, 33, 35, 50.

The sighted-user reference retains primary-eligible traces with nonempty eight-class tokens. The 43 matched tasks contain 339 eligible human trajectories. The own-success task sets contain 191, 159, and 151 reference trajectories for \cogent, \cua, and \cuauser; shared success contains 111. Systems share human reference trajectories on overlapping tasks.

\begin{table}[htbp]
\centering
\caption{Semantic-action sequence lengths in each agent's successful cohort. Means weight tasks equally after averaging humans within task; brackets are 95\% task-bootstrap intervals.}
\label{tab:a11y-lengths}
\small\begin{tabular}{@{}lrrcc@{}}
\toprule
Agent & Tasks & Humans & Agent length & Human length \\
\midrule
COGENT & 24 & 191 & 16.46 & 22.84 \\
 & & & [12.96, 19.92] & [19.04, 26.90] \\
CUA & 20 & 159 & 10.80 & 23.19 \\
 & & & [7.45, 14.35] & [18.93, 27.77] \\
CUA-User & 19 & 151 & 9.84 & 21.07 \\
 & & & [5.84, 13.95] & [16.61, 25.38] \\
\bottomrule
\end{tabular}

\end{table}

The own-success pooled distributions contain COGENT: 395 agent actions and 4,357 human actions; CUA: 216 agent actions and 3,682 human actions; CUA-User: 187 agent actions and 3,174 human actions.

Action shares normalize raw pooled counts once. Longer trajectories and tasks with more human observations therefore contribute more to these distribution estimates. Whole-task bootstrapping preserves within-task dependence.

Matched-prefix edit distance ignores unmatched suffixes. A short, prematurely ended trajectory can receive a low distance while omitting later human actions. Lengths, outcomes, and budget counts complement these estimates by describing trajectory coverage and completion. The 30-step cap limits observed agent trajectories, while human references use the full recorded sequence.

In the own-success cohort, 5/24, 4/20, 5/19 episodes reach 30 reported steps for \cogent, \cua, and \cuauser respectively; 5, 4, 5 of these budget-reaching episodes are verified successes. In the 43-task matched cohort, 16/43, 16/43, 20/43 episodes reach 30 reported steps for \cogent, \cua, and \cuauser respectively; 5, 4, 5 of these budget-reaching episodes are verified successes.

These counts describe the 30-step observation boundary; independent review determines each task outcome.

\begin{figure}[htbp]
\centering
\includegraphics[width=\linewidth]{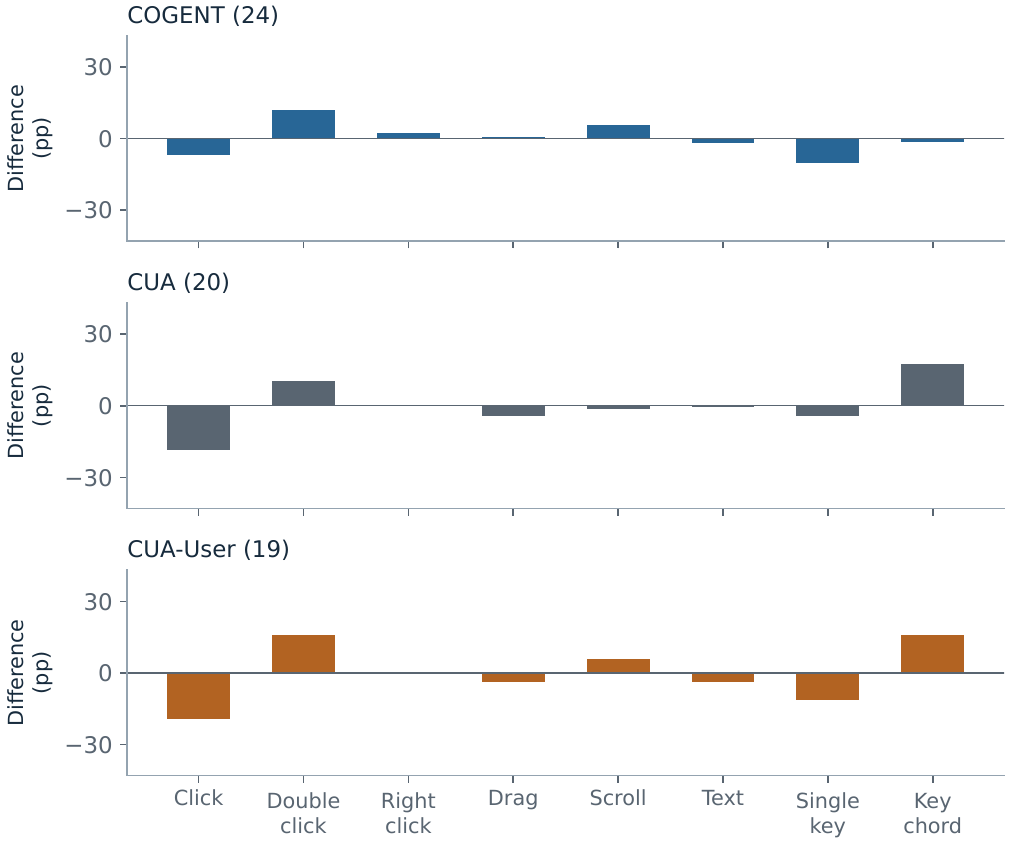}
\caption{Pooled action-share differences for each agent's successful tasks, in percentage points relative to its task-matched human pool. Each panel uses a different task set. Bars are descriptive; single key includes navigation and key chord denotes a hotkey combination.}
\label{fig:a11y-shares}
\end{figure}

\FloatBarrier
\subsection{Task-level outcomes}
\label{app:a11y-ledger}
\begin{longtable}{@{}rlllr@{}}
\caption{Outcomes on the 43 selected tasks. S: verified success; F: verified failure; U: unverifiable. Human $n$ counts sighted-user reference trajectories.}\label{tab:a11y-task-ledger}\\
\toprule
Task & COGENT & CUA & CUA-User & Human $n$ \\
\midrule\endfirsthead
\toprule Task & COGENT & CUA & CUA-User & Human $n$ \\ \midrule\endhead
1 & S & S & F & 8 \\
2 & F & S & S & 8 \\
6 & F & F & F & 8 \\
7 & S & S & S & 8 \\
9 & S & S & F & 8 \\
10 & S & S & S & 8 \\
11 & S & F & F & 8 \\
14 & F & F & F & 8 \\
18 & S & F & S & 8 \\
20 & S & F & S & 8 \\
21 & S & S & S & 8 \\
22 & S & S & S & 8 \\
24 & S & S & S & 8 \\
25 & S & S & S & 8 \\
26 & S & S & S & 8 \\
27 & S & S & S & 8 \\
28 & S & S & S & 8 \\
29 & S & S & S & 8 \\
30 & S & S & S & 8 \\
31 & S & S & S & 8 \\
32 & S & S & S & 8 \\
33 & S & S & S & 7 \\
34 & S & S & F & 8 \\
35 & F & F & S & 8 \\
36 & F & F & F & 8 \\
37 & F & F & F & 8 \\
39 & F & F & F & 8 \\
40 & F & F & F & 8 \\
41 & F & F & F & 8 \\
42 & S & F & F & 8 \\
43 & S & F & F & 8 \\
44 & F & F & F & 8 \\
45 & F & F & F & 6 \\
46 & F & F & F & 8 \\
47 & F & F & F & 8 \\
49 & U & U & F & 8 \\
50 & U & S & S & 8 \\
52 & S & F & F & 8 \\
54 & S & F & F & 8 \\
55 & F & F & F & 7 \\
57 & U & U & U & 8 \\
59 & F & F & F & 7 \\
60 & U & S & U & 8 \\
\bottomrule
\end{longtable}

\section{SENSE-42 Protocol and Supplementary Analyses}
\label{app:sense}

{
\subsection{Sensitivity to the number of SENSE-42 blocks}
We ran one simulated participant on the task sequence of human participant~001 and stopped after 11 blocks as an exploratory study, by which point the agent had completed 11 repetitions of each of the five substantive task types (55 tasks and 750 state updates). A sensitivity analysis of the 42 human participants supports this budget for an exploratory comparison. When we truncated each human session to its first 11 blocks and recomputed the cohort-level estimates, the ordering of the 21 attribute correlations closely matched the full-session ordering ($\rho_s=0.940$). Thus, the broad human pattern was already visible at this point, although agreement between an individual's first 11 blocks and their own full session was lower (median $\rho_s=0.688$). The single simulated trajectory showed agreement with the full human reference both across updates ($\rho_s=0.860$) and across task means ($\rho_s=0.830$).
}

\subsection{Task sequence and attributes}
\label{app:sense-protocol}

The simulation follows the reconstructed participant-001 sequence from SENSE-42. The full sequence contains 250 high-level activations: 45 each of Mail, file dragging, file opening, Trash, and Notes, plus 25 browser/checkpoint activations. A dragging container includes 20 inner trials and an opening container includes 30, nested within their respective high-level tasks. We analyze a partial run containing 11 repetitions of each substantive type, for 55 tasks. Browser/checkpoint activations are excluded from the substantive task-mean attribute analysis.

The seven modeled attributes are sleepiness, mental demand, temporal demand, perceived performance, effort, frustration, and attentiveness. Their meanings were generated from questionnaire item definitions and component contracts and held fixed during the run.

The agent maintains cognitive state across task boundaries. The objective and task-local buffers are reset while dynamic states and memory persist. Onboarding includes visible calibration and nine memory records representing the interface conventions taught to the user, including semantic knowledge and an executable procedure. Motor perturbation is disabled.

Sleepiness uses a 1--9 response scale; the other six questionnaire constructs use 1--7 scales. Perceived performance runs from failure to perfect, attentiveness from not at all to completely, and demand, effort, and frustration from their low to high endpoints. Internal state values use $[0,1]$ and are mapped monotonically to the questionnaire ranges. Meanings specify how each questionnaire construct can influence processing and remain fixed during the run. The ordinary appraisal process generates state updates from the agent's task experience.

\subsection{Attribute structure}
\label{app:sense-structure}

For each attribute pair, the human statistic is the median of 42 within-participant Spearman correlations. For the simulator, coefficients are calculated over 750 updates and over 55 task-mean states. The dense comparison gives $\rho_s=0.860$, MAE $=0.397$, and exact attribute-label permutation $p=0.000397$; the task-mean comparison gives $\rho_s=0.830$. The 21 edges share seven attribute nodes, motivating joint attribute-label permutation. Repeated updates within tasks are also dependent. The task-mean analysis summarizes the same single run at the task level.

Human relationship estimates use 2,000 participant-bootstrap samples; task-mean simulator relationships use 2,000 task-bootstrap samples. The two procedures preserve the sampling unit of each source; simulator intervals describe task variation within the single run.

\subsection{Attribute--behavior hypotheses}
\label{app:sense-behavior}

The temporal-demand hypothesis uses reduced pointer--keyboard switching as an operational proxy for reduced exploration under time pressure \citep{Wu2022-sj}. We operationalize exploration as pointer--keyboard switching. Mental-demand and effort hypotheses likewise specify measurable links between these constructs and computer interaction \citep{otto2013curse,brown2022humans,hart1988development}. These operational definitions determine the behavioral relationships tested below.

For temporal and mental demand, the human analysis uses 701 Mail and Notes task--questionnaire pairs, where both modalities and relevant variation are available. Effort uses all 1,078 eligible pairs. Human standardized associations remove participant and task-type additive effects, with 5,000 whole-participant bootstrap samples. The temporal-demand agent estimate uses 22 Mail and Notes tasks; the effort estimate uses all 55. Agent estimates remove task-type effects and use 5,000 task-type-stratified task bootstrap samples. Both sources use familywise 98.33\% intervals for the three-hypothesis family. Human intervals describe participant variation, and agent intervals describe task variation within the simulation.

\begin{table}[htbp]
\centering
\caption{\textbf{SENSE-42: attribute--behavior correspondence.} Bold human coefficients pass the human-support criterion; both tested agent intervals span zero. Intervals are familywise 98.33\% bootstrap intervals. A dash marks an agent hypothesis excluded by the human-support criterion.}
\label{tab:sense-behavior}
\small
\begin{tabular}{@{}p{.38\linewidth}cc@{}}
\toprule
Hypothesis & Human $\beta$ & Agent $\beta$ \\
\midrule
Temporal demand: less modality switching & $\mathbf{-0.135}$ & $+0.078$ \\
 & $[-0.223,-0.044]$ & $[-0.447,+0.545]$ \\
Mental demand: less modality switching & $+0.001$ & --- \\
 & $[-0.083,+0.092]$ & \\
Effort: more decisions per scored unit & $\mathbf{+0.110}$ & $+0.137$ \\
 & $[+0.013,+0.200]$ & $[-0.044,+0.352]$ \\
\bottomrule
\end{tabular}
\end{table}

Effort's nominal, post-hoc 90\% agent interval is $[+0.013,+0.275]$, while its familywise interval spans zero. A progress-adjusted temporal-demand sensitivity changes the human and agent point estimates to $-0.120$ and $-0.056$ respectively. The agent's estimated direction therefore depends on specification. Hypotheses were specified after inspecting earlier results, making these analyses exploratory.

\section{Data and Reproducibility}
\label{app:provenance}

The analysis materials include task selections, normalized action sequences, per-task metrics, confidence intervals, and paired comparisons. Configuration records and source checksums identify the data underlying each table and figure. Reproduction instructions specify the runtime dependencies and commands for normalization, analysis, and figure generation.

AgentNetBench results use recorded predictions and human action annotations. A11y-CUA results use executed agent commands, independently assessed task outcomes, and normalized human reference trajectories. The module-edit analysis uses scored decisions indexed by task, target, replicate, and condition. SENSE-42 results use the human and simulator attribute analyses described in Appendix~\ref{app:sense}; the simulator evidence consists of one partial run. Dataset citations identify the sources of the human observations.

Demonstration materials preserve the initial interface, intervention configuration, intermediate module records, and executed actions. Figure-generation code specifies screenshot crops and derives motor progress from the cursor observations. These materials support inspection of the implemented processing sequence.

\subsection{Agent Prompts and Interfaces}
\label{app:agent-comparison}

All experiments except CUA-User-CB use GPT-5.4 under the Azure deployment \texttt{gpt-5.4} through the Azure OpenAI Responses API (\texttt{/openai/v1/}). CUA-User-CB uses GPT-5.4 through a non-Azure API.
We detail the full A11y-CUA baseline instructions and selected AgentNetBench and \cogent prompt excerpts below, alongside module connections and intermediate representations. Quoted instructions are verbatim; template fields and interface descriptions are compacted for space. 

\paragraph{Model calls and interaction budgets.}
An interaction step is not a model call. Direct Computer-tool use requires an initial request and a continuation supplying the screenshot, with occasional replays or format repair. \cogent additionally invokes its cognitive and control processes, candidate generation, selection, and grounding as applicable; empty-memory retrieval, appraisal without dynamic states, sole-candidate selection, and keyboard operations skip calls. Equal prediction targets or A11y's common 30-step budget therefore do not imply equal model-call or token expenditure, and the behavioral comparisons with \cua and \cuauser should not be read as compute-matched tests of the architecture alone.

\paragraph{A11y-CUA: \cua.}
\begin{quote}\small
You are a computer-use agent completing a live Windows desktop task.

Application: \texttt{\{application\}}\\
Task: \texttt{\{instruction\}}

You will receive the current screenshot through the Computer tool. Take exactly one immediate UI action per turn. Do not return action JSON or prose. When the task is complete, return the text \texttt{TASK\_COMPLETE} without another computer action. Do not perform unrelated actions.

Recent actions: \texttt{\{history\}}
\end{quote}

\paragraph{A11y-CUA: \cuauser.}
\begin{quote}\small
You are operating the Windows desktop while simulating a normal human user with the supplied context. Act consistently with that context, but use the Computer tool directly rather than exposing cognitive reasoning.

Task: \texttt{\{instruction\}}; Session context: \texttt{\{session\_context\}}\\
Persona experience: \texttt{\{persona\_experience\}}; User traits: \texttt{\{traits\}}\\
Current user state: \texttt{\{states\}}; Authored prior memory: \texttt{\{memory\}}

Take exactly one immediate UI action per turn. Do not return action JSON or prose. When the task is complete, return \texttt{TASK\_COMPLETE} without another computer action. Do not perform unrelated actions.

Recent actions: \texttt{\{history\}}
\end{quote}

\paragraph{AgentNetBench: shared direct-prediction template.}
Both direct baselines use the following excerpts. Only \cuauser receives \texttt{Participant context: \{participant\_context\}}; \cogent receives the same participant context.
\begin{quote}\small
You are predicting one recorded human's next computer action, not executing a task yourself. Do not correct the recorded person's earlier choices.

Task: \texttt{\{task instruction\}}\\
Recent recorded-human actions corresponding to the prior screenshots: \texttt{\{history\}}

The immutable current screenshot is not attached. First request the current screenshot with one Computer screenshot action. It will be supplied only as \texttt{computer\_call\_output}.

After seeing the current screenshot, emit exactly the first immediate action this same human is most likely to take. Return only one semantic action in one \texttt{computer\_call} and no prose. If this human would declare the task finished successfully now, return \texttt{TASK\_COMPLETE} without a \texttt{computer\_call}. If this human would stop because the task failed, return \texttt{TASK\_FAILED} without a \texttt{computer\_call}. The action will not be executed.
\end{quote}
History contains up to two preceding screenshots and recorded-human actions. The adapter permits one current-screenshot replay and one format-repair attempt.

Participant context (\cogent and \cuauser only):
\begin{quote}\small
This trajectory was recorded from one human annotator using the annotator's personal computer. The AgentNet collection recruited annotators from internal students, external university students, annotation companies, and Prolific. The released task does not identify which source or any individual demographic traits. Predict this same recorded person's next action without inventing a profile.
\end{quote}

Direct flow: instruction/history $\rightarrow$ Computer screenshot request $\rightarrow$ current frame $\rightarrow$ action; next frame from execution (A11y) or the human recording (AgentNetBench).

\paragraph{COGENT: connected processes and passed records.}
Perception $\rightarrow$ Retrieval $\rightarrow$ Comprehension $\rightarrow$ Monitor $\rightarrow$ Appraise $\rightarrow$ Executive $\rightarrow$ candidate generation $\rightarrow$ validation/selection $\rightarrow$ grounding/execution. Completion/meta-action stops selection. The next frame feeds Comprehension; assessed actions feed lagged learning; rest triggers consolidation.

{\small
\setlength{\tabcolsep}{4pt}
\renewcommand{\arraystretch}{1.03}
\begin{longtable}{@{}>{\raggedright\arraybackslash}p{0.16\linewidth}>{\raggedright\arraybackslash}p{0.37\linewidth}>{\raggedright\arraybackslash}p{0.43\linewidth}@{}}
\toprule
Process & Prompt excerpt & Inputs $\rightarrow$ outputs \\
\midrule
\endhead
Perception &
``Approximate what THIS user actually NOTICES on the screen right now.'' &
Raw frame, session, working memory, attributes, previous expectation $\rightarrow$ percept, uncertainty, P\# motor regions and O\# state cues with text and normalized boxes; host-masked image. \\
\addlinespace
Retrieval &
``Return only existing items that actually come to mind.'' &
Session, percept/P\#/O\#, working memory, attributes, long-term store $\rightarrow$ recalled item IDs, content, and type (fact/procedure/gist). \\
\addlinespace
Comprehension &
``State the situation model in 1-2 sentences (their subjective read, possibly wrong).'' &
Session, attributes, masked image, percept/P\#/O\#, recall, working memory, previous action/expectation $\rightarrow$ situation, previous-action assessment and evidence references, working-memory writes/evictions. \\
\addlinespace
Monitor &
``Determine whether the complete SESSION OBJECTIVE has been achieved.'' &
Session, percept, uncertainty, situation, action assessment, working memory $\rightarrow$ completion Boolean and evidence. \\
\addlinespace
Appraise &
``Return one value for every listed dynamic-state key, including values that remain unchanged.'' &
All traits/states, Monitor result, percept/situation, recall, working memory, action experience $\rightarrow$ updated states and optional experience summary. \\
\addlinespace
Executive &
``Choose at most one meta-action from: ask\_help, rest, give\_up. Use null when the user continues acting.'' &
Session, relevant updated attributes, situation, Monitor result, working memory, action experience $\rightarrow$ intention, new/keep/switch/refine event, meta-action, reason. \\
\addlinespace
Candidates &
``Generate up to \{lane\_limit\} concrete, atomic motor intentions from this source lane.'' &
Session/screen ID, situation, masked image, percept/P\#/O\#, working memory, attributes, lane memory $\rightarrow$ motor intentions, support references, expected outcomes, confidence, guidance references. \\
\addlinespace
Memory lane &
``The host owns each recalled procedure's exact motor\_template.'' &
Recalled executable procedures plus candidate-generation context $\rightarrow$ procedure references and P\# target bindings; host preserves the stored operation. \\
\addlinespace
Selection &
``Select exactly one candidate by ID. Prefer an applicable memory candidate over interface or exploration.'' &
Validated memory/interface/exploration pool, session, situation, percept, masked image, working memory, attributes $\rightarrow$ candidate ID, override reason, selection reason. \\
\addlinespace
Grounding &
``Use the computer tool only to ground this exact operation on the current screenshot.'' &
Fixed motor intent and operating procedure, frame dimensions, then frame $\rightarrow$ one grounded action or \texttt{TARGET\_NOT\_FOUND}; host executes the exact command. \\
\addlinespace
Consolidation &
``GIST'', ``SCHEMATIZE'', ``COMPILE'' &
Relevant attributes, recent steps, earlier gists, frequently used facts $\rightarrow$ gist, schematized facts, compiled procedures. \\
\bottomrule
\end{longtable}
}

Session contains objective, context, and intention. Attribute selector: ``Select exactly the attributes whose meanings could materially affect how this gate performs its objective.'' Keys/meanings $\rightarrow$ selected keys $\rightarrow$ values supplied to the process. Monitor receives none; Appraise receives all. Visual verifier: ``Read the pixels, not sentence conventions or prior text hypotheses.'' Isolated source-text crop $\rightarrow$ transcription/legibility; host comparison can trigger one reselection from the same pool. Keyboard/text operations bypass visual grounding.

\end{document}